\documentclass[10pt,twocolumn,letterpaper]{article}

\usepackage{iccv}
\usepackage{times}
\usepackage{epsfig}
\usepackage{graphicx}
\usepackage{amsmath}
\usepackage{amssymb}

\usepackage{microtype}
\usepackage{epsfig}
\usepackage[table,xcdraw]{xcolor}
\usepackage{caption}
\usepackage{float}
\usepackage{placeins}
\usepackage{color, colortbl}
\usepackage{stfloats}
\usepackage{enumitem}
\usepackage{tabularx}
\usepackage{xstring}
\usepackage{multirow}
\usepackage{xspace}
\usepackage{url}
\usepackage{subcaption}
\usepackage{xcolor}
\usepackage[hang,flushmargin]{footmisc}
\usepackage{pifont}

\usepackage{mathtools}
\usepackage{float}

\usepackage[breaklinks=true,bookmarks=false]{hyperref}

\usepackage[capitalise]{cleveref}
\usepackage{soul}
\crefname{section}{Sec.}{Secs.}
\crefname{table}{Table}{Tables}
\crefname{figure}{Fig.}{Figs.}

\newcommand{\mypara}[1]{\vspace{5pt}\noindent{\bf{#1}}}
\newcommand{\Cls}{W}
\newcommand{\cls}{w}
\newcommand{\ours}{ICIS} 
\newcommand{\oursFull}{Image-free Classifier Injection with Semantics}

\newcommand{\setting}{I-ZSL}
\newcommand{\settingfull}{image-free ZSL}
\newcommand{\SettingFull}{Image-free ZSL}

\iccvfinalcopy % *** Uncomment this line for the final submission

\ificcvfinal\fi

\begin{document}

%%%%%%%%% TITLE
\title{Image-free Classifier Injection for Zero-Shot Classification}

\author{Anders Christensen\textsuperscript{1,2} \qquad 
    Massimiliano Mancini\textsuperscript{3} \qquad
    A. Sophia Koepke\textsuperscript{1} \\
    Ole Winther\textsuperscript{2,4,5,6} \qquad
    Zeynep Akata\textsuperscript{1,7} \\
    {\fontsize{10}{12}\selectfont \textsuperscript{1}University of Tübingen \qquad \textsuperscript{2}Technical University of Denmark \qquad 
    \textsuperscript{3}University of Trento}\\
    {\fontsize{10}{12}\selectfont \textsuperscript{4}University of Copenhagen \qquad
     \textsuperscript{5}Copenhagen University Hospital \qquad
     \textsuperscript{6}FindZebra \qquad
    \textsuperscript{7}MPI for Intelligent Systems} \\
    {\tt{\fontsize{10}{12}\selectfont  andchri@dtu.dk}}}

\maketitle
% Remove page # from the first page of camera-ready.
\ificcvfinal\thispagestyle{empty}\fi

%%%%%%%%% ABSTRACT
\begin{abstract}
Zero-shot learning models achieve remarkable results on image classification for samples from classes that were not seen during training. 
However, such models must be trained from scratch with specialised methods: therefore, access to a training dataset is required when the need for zero-shot classification arises. In this paper, we aim to equip pre-trained models with zero-shot classification capabilities without the use of image data. We achieve this with our proposed \oursFull\ (\ours) that injects classifiers for new, unseen classes into pre-trained classification models in a post-hoc fashion without relying on image data. Instead, the existing classifier weights and simple class-wise descriptors, such as class names or attributes, are used.  \ours\ has two encoder-decoder networks that learn to reconstruct classifier weights from descriptors (and vice versa), exploiting \mbox{(cross-)}reconstruction and cosine losses to regularise the decoding process. Notably, \ours\ can be cheaply trained and applied directly on top of pre-trained classification models. 
Experiments on benchmark ZSL datasets show that \ours\ produces unseen classifier weights that achieve strong (generalised) zero-shot classification performance.
Code is available at \href{https://github.com/ExplainableML/ImageFreeZSL}{https://github.com/ExplainableML/ImageFreeZSL}.
\end{abstract}

%%%%%%%%% BODY TEXT
\section{Introduction}
\label{sec:intro}
\begin{figure}[t!]
\centering
  \centering
  \includegraphics[width=1.\linewidth]{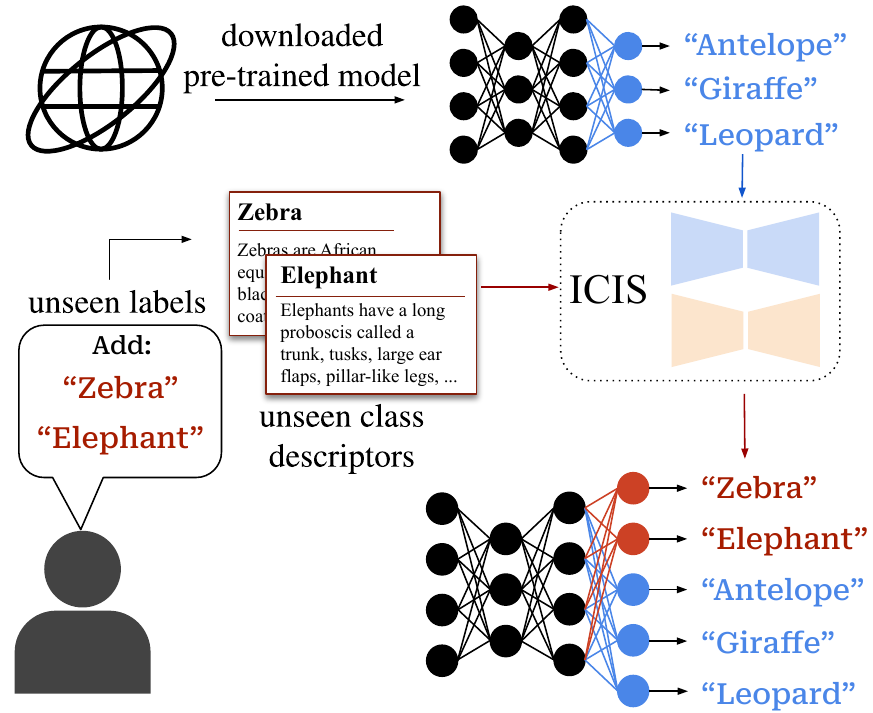}
 \caption{We propose \oursFull\ (\ours), a method that adds new classes to a pre-trained model \textit{without} access to \textit{any} image for seen and unseen classes, learning \textit{only} from class-specific descriptors (\eg attributes) and their corresponding classifier weights.
}
\label{fig:teaser}
\end{figure}
With the immense growth of user-generated data, object categories are routinely discovered or re-defined, and requirements for models change constantly. Hence, pre-trained visual classifiers can rapidly turn inadequate as their output space is limited to classes seen during training. Updating a decision maker commonly requires expensive data collection and annotation, which can be unrealistic in some domains. 

A cheaper and more appealing strategy is to use a zero-shot learning (ZSL) model \cite{Akata2016LabelEmbeddingFI,xianGoodBadUgly1,xianGoodBadUgly2,Xian2018FeatureGN} that exploits side information (\eg class label word embeddings) to recognise unseen classes in the specific domain of interest. However, i) standard classification models are not designed with zero-shot capabilities \cite{He2016DeepRL}, and ii) state-of-the-art ZSL methods are not applicable post training as one cannot regress descriptors from images \cite{LTDUO,Akata2016LabelEmbeddingFI}, train a feature generator \cite{Xian2018FeatureGN}, or learn visual-semantic embeddings \cite{xu2020attribute,xian2019semantic} without images from seen classes. Crucially, it can be infeasible for a practitioner to (re-)train a ZSL model due to computational costs or data storage issues, and some visual data might not be available for a particular task or user. 

Since a large variety of pre-trained deep learning models are readily available online, we pose the question: given a specific image classification task and a pre-trained model, can we extend it to desired but missing categories \textit{without} using images from seen or unseen classes? 
We name this task \SettingFull\ (\setting), where the aim is to categorise samples from unseen classes without using image data, by injecting new classification weights into pre-trained classification models in a post-hoc manner. We note that while vision-language models (\eg CLIP~\cite{radford2021learning}) can perform zero-shot transfer, \setting\ goes into an orthogonal direction by targeting models that are specialised for specific tasks, such as fine-grained recognition, for which model updates are needed when introducing new categories. In principle, an I-ZSL method could be combined with CLIP, \eg by estimating class-specific prompts~\cite{Parisot2023LearningTN} for unseen classes. 

We address the \setting\ task by proposing \oursFull\ (\ours), a method that directly relates semantic class descriptors (\eg class names or attributes) to the classifier weights of a pre-trained model. 
Concretely, \ours\ receives as input classifier weights of seen classes and class descriptors. It learns two encoder-decoder networks, one for descriptor vectors and one for the classifier weights. 
These networks inject priors from the respective data sources into the simple descriptor-to-weights mapping. Moreover, we supervise the encoder representations by mapping across spaces (\ie descriptor-to-weights and vice versa), using pairs of seen class descriptors and their corresponding classifier weights, in order to encourage better generalisation on unseen descriptors. 
Notably, \ours\ is trained and used for classifier injection in a post-hoc manner, \ie after the initial classification model has been trained and without access to the initial training data. We validate our approach on standard ZSL benchmark (\ie CUB~\cite{CUBdata}, SUN~\cite{SUNdata}, AWA2~\cite{xianGoodBadUgly2}), against applicable ZSL methods and adapted state-of-the-art zero- and few-shot learning approaches, consistently outperforming them in both generalised and standard evaluation.

To summarise, we make the following contributions: 
(1)~We tackle the new \settingfull\ task, where zero-shot classification is performed by injecting new classification weights into a pre-trained classification model, without access to any samples this model was trained on.
(2)~We propose \ours, a simple framework with two encoder-decoder architectures that directly predicts new classifier weights for unseen classes from limited training data, \ie classifier weights for the seen classes and corresponding class-wise descriptors.     
(3)~Despite the additional restrictions of the \setting\ setting, \ours\ achieves strong (Generalised)-I-ZSL performance on a variety of ZSL benchmarks, outperforming several zero- and few-shot approaches adapted to \setting. 

\section{Related Work}
\label{sec:related}
\mypara{Zero-Shot Learning.} aims to classify samples from classes that were not part of the original training set~\cite{LTDUO}, \eg by using class-specific semantic information to relate seen and unseen classes. 
Commonly, methods
learn to map images to class-level semantic descriptors \cite{LTDUO,Akata2016LabelEmbeddingFI,xian2016latent,romera2015embarrassingly}, improve visual-semantic embeddings \cite{xian2019semantic,GZSL-CalNet,Zhang2017LearningAD,chen2022msdn,min2020domain}, \eg using part-aware training strategies \cite{xu2020attribute,xie2020region}, or train generative models to synthesise visual features for unseen classes~\cite{Xian2018FeatureGN,XianCVPR2019a,CADA-VAE,Zhu2018AGA,Zhu2019LearningFT,kong2022compactness,han2021contrastive,keshari2020generalized,chen2021free}. Recent methods have also considered novel ways of comparing visual features and semantic descriptions using visual patches \cite{xu2022vgse} or noisy documents \cite{naeem2022i2dformer}, and descriptions generated with GPT-3~\cite{ferjad2022i2mvformer,brown2020language}.  
The aforementioned methods assume full access to the training data and the possibility to train a ZSL model from scratch. In this work, we remove this assumption and create a zero-shot classification model from a pre-trained classification one, \textit{without} access to the images it was trained on, using only semantic descriptions of the classes. 

\mypara{Generating classifiers for ZSL.} \ours\ addresses the lack of training images in a direct way, learning a function that maps class descriptors to their corresponding classifier weights. In this regard, \ours\ has some similarities to hypernetworks~\cite{HyperNetworksHa} and previous works that infer weights for zero-shot classification. Most of these rely on graph-convolutional neural networks (GCNs) \cite{kipf2017semi}: \cite{wang2018zero,kampffmeyer2019rethinking} predicts classifier weights given semantic descriptions from a class-hierarchy as side information. \cite{wang2018zero, kampffmeyer2019rethinking, Gidaris2019GeneratingCW} use seen class images to refine the graph, fine-tune input features, and train a denoising autoencoder, respectively. Other approaches 
learn virtual classifiers for unseen classes~\cite{changpinyo2016synthesized}  or sample images to train the classifier generator~\cite{Li2019RethinkingZL} when training the base classification model. Different from these approaches, \ours\ does not use the original training data or scarcely available auxiliary information (\eg class hierarchies). 
To the best of our knowledge, among previously published ZSL literature, only ConSE~\cite{Norouzi2014ZeroShotLB} and COSTA~\cite{mensink2014costa} can directly perform zero-shot classification without relying on images. While ConSE~\cite{Norouzi2014ZeroShotLB} maps images to the attribute space by a weighted combination of the seen class embeddings, COSTA~\cite{mensink2014costa} directly estimates classifier weights for unseen classes by exploiting their co-occurrence/similarity with seen ones.  Despite the appealing simplicity of ConSE and COSTA, they do not explicitly consider the image-free setting, although being applicable for it. 

\mypara{Incremental, few-shot learning.}
Since \ours\ extends a classification network to new classes, it is related to incremental \cite{li2017learning,Kirkpatrick2017OvercomingCF,rebuffi2017icarl,douillard2020podnet} and continual learning \cite{zenke2017continual,lopez2017gradient,prabhu2020gdumb,de2021continual}, and in particular few-shot incremental learning (IL) \cite{Zhang2021FewShotIL,dong2021few,tao2020few,akyurek2021subspace,cheraghian2021semanticfsl,hersche2022constrained}. Interestingly, previous works also merged ZSL and IL~\cite{xue2017incremental,wei2021incremental1,wei2021incremental2,kankuekul2012online} to either improve ZSL models with a stream of data containing also unseen class images \cite{xue2017incremental,wei2021incremental1}, learn new attributes \cite{kankuekul2012online}, or increase IL performance \cite{wei2021incremental2}. However, different from these works, we add new classes to the model without access to any image data of new or old classes. 
This makes our model insusceptible to catastrophic forgetting \cite{french1999catastrophic,Kirkpatrick2017OvercomingCF}, a common issue in IL. 

Considering few-shot IL, our model is most closely related to \cite{akyurek2021subspace}, which explores different ways to predict classifier weights for new classes by using few-shot samples, side information, and subspace projections. Differently from \cite{akyurek2021subspace}, we have no images informing our descriptor-to-weights mapping and we do not rely on subspace regularisers. Instead, our \ours\ learns to encode class descriptors and classifier weights into a shared embedding space whilst preserving their respective structures via reconstruction objectives. 

\section{Methodology}
\label{sec:method}
\begin{figure*}
\centering
  \centering
  \includegraphics[width=1.\textwidth]{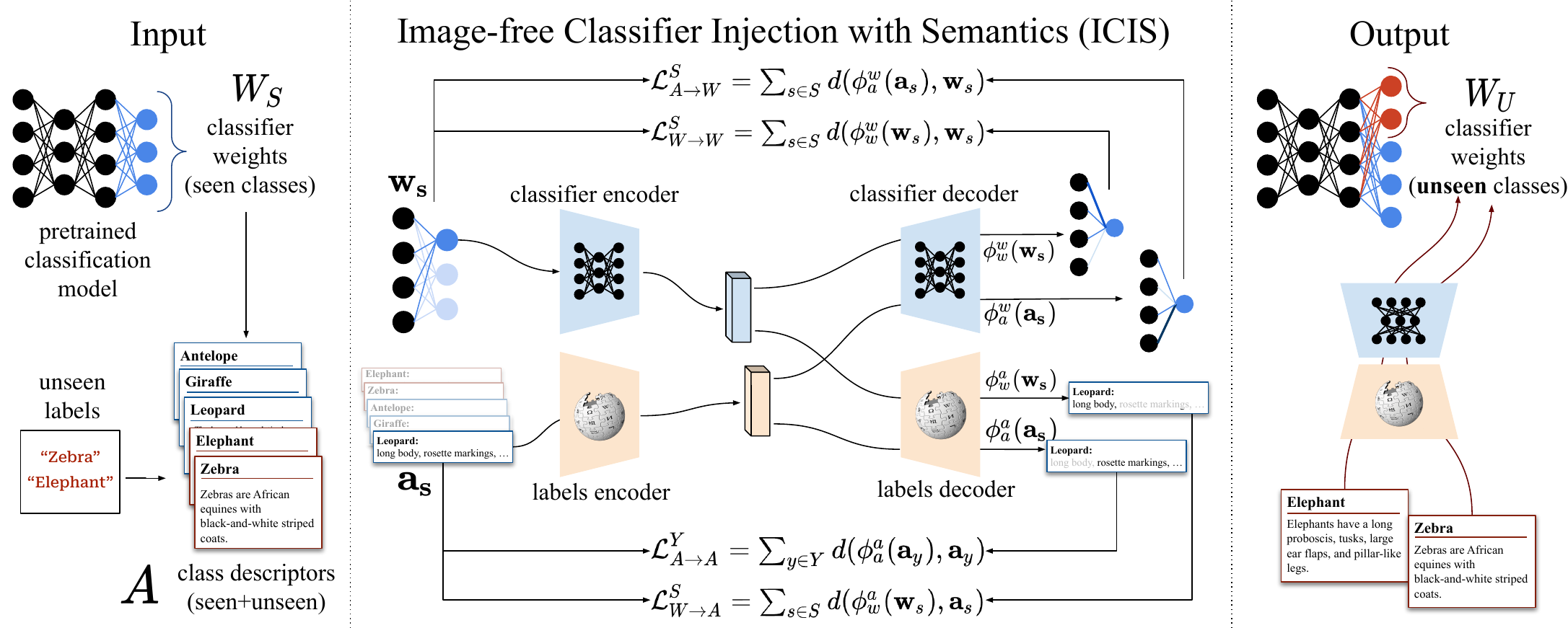}
\caption{\textbf{\oursFull\ (\ours)}. Given a pre-trained model with classifier weights $\Cls_S$ trained on the set of seen classes $S$, our goal is to inject new classifier weights with weights for unseen classes (\eg red nodes) into the set of existing weights $\Cls_S$ \textit{without any images}, using only class-specific descriptors. To tackle this task, we instantiate two encoder-decoder architectures (blue and ocher) that map inputs within and across the two spaces, supervising them using pairs of descriptors and classifier weights for seen classes. During inference, we use the descriptor encoder and the weights decoder to infer weights for an unseen class given its descriptor. The inferred weights can then be injected into the set $\Cls_S$ (right), allowing the pre-trained model to recognise images of unseen classes.}
 
\label{fig:method}
\end{figure*}

We first formalise the \settingfull\ task in Section \ref{sec:problem-formulation}. We then describe our method, \ours, a simple, yet effective approach for \setting\ (Section \ref{sec:ours-method}). 
Figure~\ref{fig:method} provides an overview of our task and approach.

\subsection{The \SettingFull\ task}
\label{sec:problem-formulation}
In this work, we tackle the task of adapting a pre-trained classification model to categorise images of previously unseen classes, without access to any images. 

Let us denote a given  classification model as $\Phi:  \mathcal{X} \rightarrow S$, where $\mathcal{X}$ is the image space, and $S$ is the set of seen categories. Without loss of generality, we assume that $\Phi$ is composed of two modules, \ie a feature extractor $F$ and a classifier with weights $\Cls_S$. We assume that $\Phi$ is pre-trained on an unavailable dataset with annotated seen class images. Our objective is to inject new classifiers into $\Phi$ to be able to classify images from a given set of target classes $Y$, \ie producing the extended model $\hat{\Phi}:\mathcal{X} \rightarrow Y$. Note that we can have $Y\cap S=\emptyset$, as in the ordinary ZSL setup, or $Y\cap S=S$ if we wish to continue recognising seen classes, as in the more challenging case of GZSL. In the following, we will denote the set of unseen classes in $Y$ as $U$, \ie $U=Y\setminus S$. 

To inject classifiers that extends $\Phi$ to $U$, we assume to have a set of vectors $A$, which describe the classes in $S$ and $U$. We denote the semantic descriptors for classes in $S$ and $U$ as $A_S=\{\mathbf{a}_s\}_{s\in S}$ and $A_U=\{\mathbf{a}_u\}_{u\in U}$, with $A=A_S \cup A_U$. The only constraint for $A$ is that all descriptors come from the same source (\eg word embeddings, expert annotations), allowing us to model the relationships between seen and unseen classes. In principle, the set of descriptors can be obtained from generic linguistic sources (\eg language models, word embeddings) and thus we can also use just embeddings of the class names as inputs.

From the above definition, the main challenge of \settingfull\ is the lack of \textit{any} training images. This constraint prevents the use of standard ZSL techniques, such as generating features for unseen classes \cite{Xian2018FeatureGN,CADA-VAE} or learning specific compatibility functions \cite{xian2019semantic,xu2020attribute}, since we cannot learn how visual inputs relate to the descriptor space. We instead work in a different space: the classification weights of the given pre-trained model. Specifically, we assume that the existing classifier weights $\Cls_S$ for the seen classes $S$ already encode visual-semantic knowledge, and act as visual descriptors of the seen classes. From this assumption, we approach the task of \settingfull\ with the intention of mapping class descriptors to their corresponding classifier weights. We can then perform zero-shot classification by equipping $\Phi$ with inferred weights for unseen classes. In the following, we describe how we implement this mapping in \ours.

\subsection{\oursFull}
\label{sec:ours-method}
The \oursFull\ (\ours) framework for the \setting\ extends a base model by predicting classifier weights from class-specific descriptors with an autoencoder structure and training objectives that regularise learning in this extremely data-scarce setting.  Here, we first describe our base model and its extensions for \setting.

\mypara{Base model.} Our goal is to produce a module mapping descriptors of unseen classes to corresponding classifier weights. Formally, let us denote the set of classifier weights of $\Phi$ for the seen classes in $S$ by $\Cls_S=\{\mathbf{\cls}_s\}_{s\in S}$. We want to learn a mapping $\phi_a^\cls: \mathcal{A} \rightarrow \mathcal{\Cls}$ where $\mathcal{A}$ is the descriptor space and $\mathcal{\Cls}$ the classifier weights space. The simplest approach to learn $\phi_a^\cls$ is by exploiting the available descriptors and classifier weights of the seen classes. With the latent space $\mathcal{Z}$, the descriptor encoder $E_a: \mathcal{A} \rightarrow \mathcal{Z}$, and the classifier weights decoder $D_\cls: \mathcal{Z} \rightarrow \mathcal{\Cls}$, we can define $\phi_a^\cls = D_\cls \circ E_a$. We learn $\phi_a^\cls$ by minimising the following regression objective:
\begin{equation}
    \label{eq:a->w}
    \mathcal{L}^{S}_{A\rightarrow \Cls} = \sum_{s\in S} d(\phi(\mathbf{a}_s),\mathbf{\cls}_s),
\end{equation}
where $d$ is a generic distance function (\eg mean-square error). During inference, we infer a classifier for an unseen target class $u \in U$ as:
\begin{equation}
    \label{eq:inference}
    \mathbf{\cls}_u = \phi_a^\cls(\mathbf{a}_u) = D_\cls(E_a(\mathbf{a}_u)),
\end{equation}
and inject $\mathbf{\cls}_u$ into the resulting classification head $\Cls$. 
If the downstream classification task is in the generalised zero-shot setting, we have $\Cls=\Cls_S\cup \Cls_U$ with $\Cls_U = \lbrace \mathbf{\cls}_u \rbrace_{u\in U}$, while for the ordinary zero-shot setting $\Cls=\Cls_U$. 

The base model achieves good results in the zero-shot setting, but its performance rapidly decreases in the generalised case. In the following, we describe how $\phi_a^\cls$ can be improved to better generalise to unseen class descriptors in $A_U$.

\mypara{Comparing vectors through cosine-similarity.} The choice of the distance function $d$ heavily affects the performance of the framework. While multiple choices are possible, we employ a simple implementation based on cosine-similarity:
\begin{equation}
\label{eq:cosine}
    d(\mathbf{v}, \mathbf{q}) =1-\cos(\mathbf{v},\mathbf{q}) =  1-\frac{\mathbf{v}\cdot \mathbf{q}}{||\mathbf{v}||\cdot ||\mathbf{q}||},
\end{equation}
where $\mathbf{v},\mathbf{q} \in \mathbb{R}^m$. Using the cosine distance, e.g. instead of L2 distance, allows the network to focus on the angular alignment of the injected weights with respect to the pre-trained ones, ignoring differences in magnitude. We found the angular alignment to promote better generalisation and to reduce the usual bias on seen classes, common in GZSL~\cite{chao2016empirical}.

\mypara{Autoencoding to preserve semantic structure.}
Both the classifier weights in $\Cls_S$ and the descriptors in $A$ are robust sources of visual/semantic information. Indeed, $\Cls_S$ has been trained on a dataset $\mathcal{T}$ with images of the seen classes, and consequently encodes distinctive visual relationships/similarities depicted in $\mathcal{T}$. At the same time, $A$ might be derived from human expertise (\eg animal descriptions) or from natural language resources trained on large corpora (\eg word embeddings, language models). Given that we face an extreme low-data scenario, it would be ideal to regularise the descriptor-to-weights mapping $\phi_a^\cls$ by preserving the structure in the spaces $\mathcal{A}$ and $\mathcal{\Cls}$. 

We found that simply adding specific autoencoder structures for each of the spaces achieves this goal. Formally, we instantiate two additional networks $D_a: \mathcal{Z} \rightarrow \mathcal{A}$ mapping vectors in $\mathcal{Z}$ back to the descriptor space $\mathcal{A}$, and $E_\cls: \mathcal{\Cls} \rightarrow \mathcal{Z}$ mapping classifier weights to the latent space $\mathcal{Z}$. We can then enforce the original structure of the descriptor space in $\phi_a^\cls$ by regularising $E_a$ and minimising the following objective:
\begin{equation}\label{eq:a->a}
    \begin{aligned}
    \mathcal{L}^S_{A\rightarrow A} &= \sum_{s \in S} d(\phi^{a}_a(\mathbf{a}_s),\mathbf{a}_s)\\
    &= \sum_{s \in S} d(D_a(E_a(\mathbf{a}_s)),\mathbf{a}_s),
\end{aligned}
\end{equation}
where $\phi^a_a = D_a \circ E_a$. Note that in case we know the target unseen classes at training time, we can minimise the objective over the full set $Y$ in Eq.~\eqref{eq:a->a}, rather than just $S$. Similarly, we enforce that $\mathcal{Z}$ additionally preserves the structure of the given classifier weights of seen classes by minimising: 
\begin{equation}\label{eq:w->w}
    \begin{aligned}
    \mathcal{L}^{S}_{\Cls\rightarrow \Cls} &=  \sum_{s \in S} d(\phi_\cls^\cls(\mathbf{\cls}_s)), \mathbf{\cls}_s)\\
    &= \sum_{s \in S} d(D_\cls(E_\cls(\mathbf{\cls}_s)), \mathbf{\cls}_s),
\end{aligned}
\end{equation} 
where $\phi^\cls_\cls = D_\cls \circ E_\cls$. With Eq.~\eqref{eq:a->a} and Eq.~\eqref{eq:w->w}, we are enforcing that vectors encoded in $\mathcal{Z}$ can be reconstructed by their space-specific decoders. This encourages $E_a$ and $D_\cls$ to not only focus on the descriptor-to-weights mapping, but also to preserve the information encoded in their respective input/output spaces $\mathcal{A}$ and $\mathcal{\Cls}$.

\mypara{Latent space alignment.}
While the autoencoders preserve structure from the respective spaces, they do not ensure that the two latent spaces are aligned, \ie a descriptor $\mathbf{a}_y$ and its corresponding classifier weights $\mathbf{\cls}_y$ may be distant in $\mathcal{Z}$. While vectors in the two spaces are already partially aligned by means of Eq.~\eqref{eq:a->w}, we can exploit the two additional modules $E_\cls$ and $D_a$ to further impose alignment via a symmetric objective of mapping classifier weights to descriptors. This is achieved with the following term: 
\begin{equation}
    \label{eq:w->a}
    \begin{aligned}
    \mathcal{L}^{S}_{\Cls\rightarrow A} &=\sum_{s\in S} d(\phi_a^\cls(\mathbf{\cls}_s), \mathbf{a}_s)\\
    &= \sum_{s\in S} d(D_a(E_\cls(\mathbf{\cls}_s)), \mathbf{a}_s),
    \end{aligned}
\end{equation}
where $\phi^a_\cls = D_a \circ E_\cls$. Combining this objective with the previous ones promotes the regularisation of $\phi_a^\cls$ toward aligned and structure-aware representations in $\mathcal{Z}$.

\mypara{Full objective.}
Merging the four terms defined in the previous sections, our final objective becomes: 
\begin{equation}
     \label{eq:final-loss}
    \mathcal{L}=\mathcal{L}^{S}_{A\rightarrow \Cls}+  \mathcal{L}^{Y}_{A\rightarrow A} +  \mathcal{L}^{S}_{\Cls\rightarrow \Cls} + \mathcal{L}^{S}_{\Cls\rightarrow A},
\end{equation}
where the first is our target mapping (Eq.~\eqref{eq:a->w}), the second and the third terms are the structure preserving ones within each of the spaces (Eq.~\eqref{eq:a->a} and Eq.~\eqref{eq:w->w}), and the last term is the reverse weights-to-descriptor objective, that encourages further alignment in $\mathcal{Z}$ (Eq.~\eqref{eq:w->a}).

\section{Experiments}\label{sec:experiments}
In this section, we provide experimental results of \ours\ in the \settingfull\ task. We first describe the datasets, baselines used in our experiments, and training details of \ours\ in \Cref{suse:datasets_baselines}. Experimental results of (generalised) zero-shot classification performed with injected weights from \ours\ are showcased in \Cref{suse:experimental_results}), and \Cref{suse:model_ablation} analyses the individual elements of our framework. 

\subsection{Experimental setting}\label{suse:datasets_baselines}
Here, we describe the three datasets used for evaluating our framework \ours\ and the baselines that we compare to.

\mypara{Datasets.}
We evaluate \setting\ performance on the standard ZSL benchmark datasets CUB \cite{CUBdata}, AWA2 \cite{xianGoodBadUgly2}, and SUN \cite{SUNdata}. We use splits of seen and unseen classes proposed in \cite{xianGoodBadUgly1,xianGoodBadUgly2} for (generalised) zero-shot learning. 
CUB \cite{CUBdata} is a dataset for fine-grained bird classification with 200 categories. Following \cite{Akata2016LabelEmbeddingFI,xianGoodBadUgly2}, we split the classes into 150 seen classes and 50 unseen classes: this means that our \setting\ training set consists of 150 pairs of descriptors and classifiers. AWA2 \cite{xianGoodBadUgly2} is a coarse-grained classification dataset 50 different animals (40 seen and 10 unseen classes). 
Finally, SUN is a dataset of scene classification consisting of 717 indoor and outdoor scenes. Following \cite{lampert2013attribute}, we consider 645 classes as seen and 72 as unseen, using the split defined for (generalised) zero-shot learning in \cite{xianGoodBadUgly2}.

\mypara{Baselines.}
To evaluate \ours, we compare with baselines and existing methods in the literature that are compatible with \cite{Norouzi2014ZeroShotLB, mensink2014costa} or adaptable to \cite{akyurek2021subspace, Gidaris2019GeneratingCW, xu2022vgse} \setting. 
\textit{ConSE} \cite{Norouzi2014ZeroShotLB} and \textit{COSTA} \cite{mensink2014costa} perform a weighted sum of existing seen class elements (embeddings the first, classifiers the latter) using as input either the current test sample (ConSE) or co-occurrence statistics (COSTA). \textit{VGSE} \cite{xu2022vgse} proposes to extract visual attributes from the seen classes, predicting the unseen class attributes via either weighted average of embedding similarities (\textit{WAvg.}) or similarity matrix optimisation (\textit{SMO}). We test these two strategies for \setting, as they do not require seen class images.

Finally, we adapt the approaches of \cite{akyurek2021subspace, Gidaris2019GeneratingCW} from the FSL literature. Subspace regularisers (\textit{Sub.~Reg.} \cite{akyurek2021subspace}) 
encourage novel classifiers to be within the subspace spanned by the base classifiers of the given model, under semantic guidance by descriptors. wDAE \cite{Gidaris2019GeneratingCW} trains a GNN-based denoising autoencoder to improve initial estimate classifiers. In both cases, we train a simple MLP (with the same structure of \ours) to provide candidate unseen class classifiers. We then apply the regularising projections on the predicted weights for Sub.~Reg., while we use them as input to DAE-GNN instead of image feature means. In the supplementary material, we show results when combining these approaches with our \ours, since the methods are complementary.

{\setlength{\tabcolsep}{4pt}
\renewcommand{\arraystretch}{1.1}
\begin{table*}[t]
\centering
\resizebox{0.89\linewidth}{!}{
\begin{tabular}{l|ccc|ccccccccc}
\multicolumn{1}{c|}{}                                                     & \multicolumn{3}{c|}{\textbf{Zero-Shot Accuracy}}                                                                                            & \multicolumn{9}{c}{\textbf{Generalised Zero-Shot Accuracy}}                                                                                                                                                                                                                                                                                                                                                                            \\
\multicolumn{1}{c|}{}                                                     & \textbf{CUB}                                 & \textbf{AWA2}                                & \textbf{SUN}                                  & \multicolumn{3}{c}{\textbf{CUB}}                                                                                                            & \multicolumn{3}{c}{\textbf{AWA2}}                                                                                                           & \multicolumn{3}{c}{\textbf{SUN}}                                                                                                           \\
\multicolumn{1}{c|}{\multirow{-3}{*}{\textbf{
\begin{tabular}[c]{@{}c@{}}Image-free \\ Zero-Shot Learning \end{tabular}}}}                    & \textbf{Acc}                                  & \textbf{Acc}                                  & \textbf{Acc}                                   & \textbf{u}                                   & \textbf{s}                                   & \multicolumn{1}{c|}{\textbf{H}}               & \textbf{u}                                   & \textbf{s}                                   & \multicolumn{1}{c|}{\textbf{H}}               & \textbf{u}                                   & \textbf{s}                                   & \textbf{H}                                   \\ \hline
ConSE  \cite{Norouzi2014ZeroShotLB}                                                        & $41.9$                                 &          $44.0$                                    &                                    $44.4$           & $0.5$                                  & $88.0$                                 & \multicolumn{1}{c|}{$0.9$}              &     $3.0$                                         &           $96.1$                                   &  \multicolumn{1}{c|}{$5.7$}                         &             $0.1$                                 &                   $47.9$                           &       $0.1$                                       \\

COSTA \cite{mensink2014costa}                                                                   &        $31.9$                                      &                   $40.9$                           &     $19.9$                                          &      $0.0$                                        &              $87.6$                                & \multicolumn{1}{c|}{$0.0$}                         &                                  $0.0$            &       $96.1$                                       & \multicolumn{1}{c|}{$0.0$}                         &    $0.0$                                          &                            $50.1$                  &                              $0.0$                \\

Sub.~Reg.* \cite{akyurek2021subspace}                                                                   &        $37.6$                                    &     $37.5$                                         &         $48.3$                                      &       $0.0$                                       &       $87.6$                                       & \multicolumn{1}{c|}{$0.0$}                         &                                    $0.0$          &                                 $96.1$             & \multicolumn{1}{c|}{$0.0$}                         &          $0.0$                                    &          $50.1$                                    &     $0.0$                                         \\
wDAE*  \cite{Gidaris2019GeneratingCW}                                                        & $38.2$                                 &          $37.0$                                    &                                    $49.9$           & $0.0$                                  & $87.3$                                 & \multicolumn{1}{c|}{$0.0$}              &     $0.1$                                         &           $96.0$                                   &  \multicolumn{1}{c|}{$0.3$}                         &             $0.0$                                 &                   $49.3$                           &       $0.0$                                       \\
WAvg*  \cite{xu2022vgse}                                                       & $2.0$                                 &          $20.1$                                    &                                    $1.4$           & $1.9$                                  & $52.3$                                 & \multicolumn{1}{c|}{$3.7$}              &     $5.5$                                         &           $92.4$                                   &  \multicolumn{1}{c|}{$10.4$}                         &             $0.0$                                 &                   $50.1$                           &       $0.0$                                       \\
SMO*  \cite{xu2022vgse}                                                        & $45.1$                                 &          $55.4$                                    &                                    $42.7$           & $39.2$                                  & $52.3$                                 & \multicolumn{1}{c|}{$44.8$}              &     $31.8$                                         &           $92.4$                                   &  \multicolumn{1}{c|}{$47.3$}                         &             $42.5$                                 &                   $1.6$                           &       $3.1$                                       \\
\ours\ (Ours)                                                                      & $\mathbf{60.6}$                                 &                $\mathbf{64.6}$                              & $\mathbf{51.8} $                                        &              $45.8$                                &   $73.7$                                           & \multicolumn{1}{c|}{$\mathbf{56.5}$}                         &                              $35.6 $                &         $93.3$                                     & \multicolumn{1}{c|}{$\mathbf{51.6}$}                         &  $45.2$                                            &                $25.6$                              &   $\mathbf{32.7}$                                           \\ 
\end{tabular}
}
\caption{Comparison between our proposed framework \ours\ and existing methods in the literature applicable or adaptable to the \settingfull\ (\setting) setting using standard ZSL benchmarks (\ie CUB, AWA2, and SUN). We measure the results as unseen accuracy (Acc) for the zero-shot task, unseen (u) and seen (s) accuracy and their harmonic mean (H) for the generalised zero-shot setting. Methods marked with * are adapted to the image-free setting. 
}
\label{table:results}
\end{table*}
}

\mypara{Hyperparameter selection without images.}
Since we have no access to any image data, we cannot select hyperparameters based on downstream classification performance on neither unseen nor seen classes. Instead, to determine suitable hyperparameters for our methods and all the baselines, we split the \textit{descriptor-weights-pairs} of the seen classes into a training and validation split, using the same validation splits of seen classes proposed in \cite{xianGoodBadUgly2,xianGoodBadUgly1} for standard (generalised) zero-shot learning, without using any image.

While providing reliable hyperparameters, we found that using the validation loss as criterion for early stopping leads to poor convergence, due to our extremely low-shot learning setup. We sidestep this problem by training on all samples, and applying a simple stopping criterion based on the slope of the training curve. We provide more details on the hyperparameters and the criterion in the supplementary. 

\mypara{Implementation details.}\label{suse:implementation_details}
As the given pre-trained classification model, we use a ResNet101~\cite{He2016DeepRL} trained to classify only the seen classes in each respective dataset, as is common in the ZSL literature \cite{XianCVPR2019a}.
Across all datasets, the encoders and decoders of the simple \ours\ framework are single layer linear mappings, with the encoder mappings being followed by a ReLU activation function. For CUB and AWA2, the hidden dimensionality is 2048, while it is 4096 for SUN. The mappings are trained with the Adam optimiser~\cite{kingma2014adam} with a learning rate of $10^{-5}$ and $(\beta_1, \beta_2)=(0.9, 0.999)$. We employ a batch size of 16 on CUB and SUN, while we set the batch size to 20 for AWA2.

\subsection{Results on \SettingFull}\label{suse:experimental_results}
In \Cref{table:results}, we show experimental results of our \ours\ and the competitors in \setting, using the standard dataset-specific attributes adopted in ZSL~\cite{xianGoodBadUgly2} with dimensionalities 312, 85, and 102 for CUB, AWA2, and SUN, respectively. This makes the results more comparable to those from ordinary ZSL approaches that have access to images. 

As the table shows, \ours\ consistently outperforms all competitors, both in standard and generalised ZS classification. When only unseen classes are predicted (zero-shot accuracy), \ours\ improves over the best ZS baselines by a large margin in all settings (\ie +15.5\% on CUB and +9.2\% on AWA2 over SMO, and +7.4\% over ConSE on SUN). The same applies to the adapted FS baselines, with \ours\ outperforming the best method by more than 20\% on both CUB and AWA2, and by 2\% on SUN. 

These margins are even more evident in the generalised setting, where \ours\ is the only method that consistently provides a good trade-off between accuracy on seen and unseen classes (\eg achieving a harmonic mean of 56.5 on CUB), while all other baselines but SMO fail in this setting. This is due to their inherent bias toward seen categories that our method largely alleviates. These results demonstrate that \ours\ can successfully inject inferred classifiers for unseen classes into an existing classification network without relying on image training data.

\begin{table*}[t]
\centering
\resizebox{0.89\linewidth}{!}{
\begin{tabular}{l|ccc|ccccccccc}
\multicolumn{1}{c|}{}                                                     & \multicolumn{3}{c|}{\textbf{Zero-Shot Accuracy}}                                                                                            & \multicolumn{9}{c}{\textbf{Generalised Zero-Shot Accuracy}}                                                                                                                                                                                                                                                                                                                                                                            \\
\multicolumn{1}{c|}{}                                                     & \textbf{CUB}                                 & \textbf{AWA2}                                & \textbf{SUN}                                  & \multicolumn{3}{c}{\textbf{CUB}}                                                                                                            & \multicolumn{3}{c}{\textbf{AWA2}}                                                                                                           & \multicolumn{3}{c}{\textbf{SUN}}                                                                                                           \\
\multicolumn{1}{c|}{\multirow{-3}{*}{\textbf{\ours\ Ablation}}}                    & \textbf{Acc}                                  & \textbf{Acc}                                  & \textbf{Acc}                                   & \textbf{u}                                   & \textbf{s}                                   & \multicolumn{1}{c|}{\textbf{H}}               & \textbf{u}                                   & \textbf{s}                                   & \multicolumn{1}{c|}{\textbf{H}}               & \textbf{u}                                   & \textbf{s}                                   & \textbf{H}                                   \\ \hline
MLP base model                                                     &          $41.4$                                    &     $46.8$                                         &         $49.7$                                      &       $0.0$                                       &       $87.6$                                       & \multicolumn{1}{c|}{$0.0$}                         &                                    $2.0$          &                                 $95.9$             & \multicolumn{1}{c|}{$4.0$}                         &          $0.0$                                    &          $50.1$                                    &     $0.0$                                         \\
+ Cosine loss                                                            & $52.7$                                 &       $50.9$                                       & $48.5$                                        &  $36.6$                                            &     $76.6$                                         &    \multicolumn{1}{c|}{$49.5$}                         &        $27.3$                                      &   $93.1$                                           &  \multicolumn{1}{c|}{$42.2$}                         &        $21.9$                                      &          $43.6$                                    &  $29.2$                                            \\
+ Within spaces                                                            & $58.0$                                 &       $63.5$                                       & $50.8$                                        &  $39.6$                                            &     $77.6$                                         &    \multicolumn{1}{c|}{$52.5$}                         &        $33.8$                                      &   $93.5$                                           &  \multicolumn{1}{c|}{$49.6$}                         &        $38.9$                                      &          $32.5$                                    &  $35.4$                                            \\
+ Across spaces                                                                    &  $60.1$                                            &      $64.8$                                        &            $51.7$                                  &    $44.5$                                          &               $75.0$                                & \multicolumn{1}{c|}{$55.9$}                         &   $35.5$                                           &     $93.6$                                         & \multicolumn{1}{c|}{$51.4$}                         &                       $44.0$                       &     $27.6$                                         &        $33.9$                                      \\

+ Include $A_U$                                                            & $60.6$                                 &       $64.6$                                       & $51.8$                                        &  $45.8$                                            &     $73.7$                                         &    \multicolumn{1}{c|}{$56.5$}                         &        $35.6$                                      &   $93.3$                                           &  \multicolumn{1}{c|}{$51.6$}                         &        $45.2$                                      &          $25.6$                                    &  $32.7$                                            \\
                                         
\end{tabular}
}
\caption{Ablation study of the individual elements of our proposed \ours. 
Starting from the MLP base model with standard L2 loss, we replace the latter with the cosine distance. We then analyse the encoder-decoder architectures by performing mappings within and across spaces. Finally, we check the impact of adding unseen class descriptors during training.} 
\label{table:ablation}
\end{table*}

\subsection{Analyses of our \ours\ framework}\label{suse:model_ablation}
In this section, we analyse the contributions of the individual components of our \ours\ framework, ablating the impact of each technical component in this extremely low-shot setting, and how the choice of embeddings and pre-training affects \ours\ performance. 

\mypara{Ablation study.}
Here, we validate the benefit of the main components of our proposed model, showing the results in \Cref{table:ablation}. We start with the MLP base model, that maps descriptors to classifier weights and it is trained with a standard L2 loss. This model achieves good results on ZS but its performance drastically decreases in GZSL, since the resulting classifiers tend to overly predict seen classes. Changing the objective to a cosine loss (+ \textit{Cosine loss}) largely mitigates this bias problem (\eg 49.5 harmonic mean on CUB, with 36.6\% unseen accuracy vs 0.0\% of the base model). The angle-based loss leaves the norm of the output weights to be implicitly regularised, but improves compatibility of the new predicted classification weights with the existing ones for seen classes, largely increasing the entropy on the prediction between the two sets (\eg from 0.89 to 2.8 on {CUB}, see the additional analysis in the supplementary).
    
With `\textit{Within spaces}' in \Cref{table:ablation}, we refer to the intermediate step of including the reconstruction losses on class descriptors (Eq. \eqref{eq:a->a}) and classifiers (Eq. \eqref{eq:w->w}). This leads to improved results on all metrics and datasets, \eg from 50.9\% to 63.5\% unseen accuracy on AWA2 for ZS, and from 29.2 to 35.4 harmonic mean on SUN GZSL. We ascribe this to the implicit regularisation added by forcing the embeddings to retain semantic knowledge, such that the class-specific distinctive attributes can again be decoded from them.

Including the reconstruction objective from weights to class descriptors Eq. \eqref{eq:w->a} leads to improved results everywhere, but on the SUN seen class accuracy. This additional alignment objective, matching latent embeddings across spaces, appears to be otherwise beneficial for the extreme low-data setting of the considered datasets. 

Finally, including descriptors of unseen classes (+ Include $A_U$) in the descriptor-to-descriptor mapping during training (Eq. \eqref{eq:a->a}) leads to marginal improvements, suggesting that while unseen class information may help, we can still train \ours\ without a-priori knowledge of the target unseen classes.

\begin{figure*}[t!]
    \centering
        \begin{subfigure}[t]{0.41\linewidth}
        \centering
        \includegraphics[width=\linewidth]{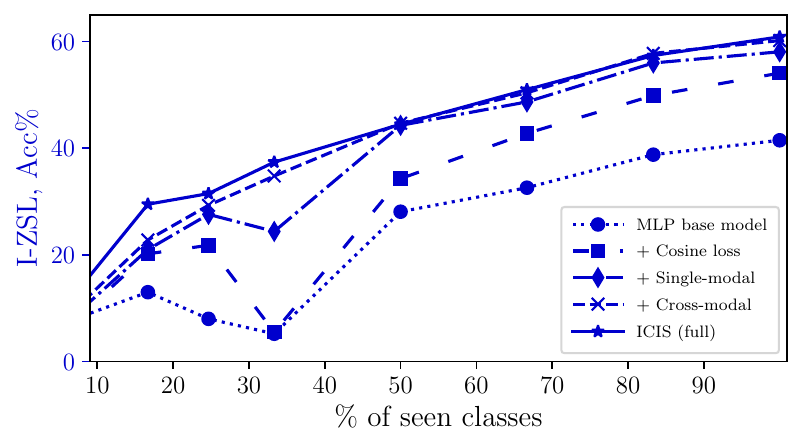}
        %\vspace{-2.0em}
        \caption{Impact on I-ZSL performance, unseen Acc\%.}
        \label{fig:ablation_ours_numsamples_zsl}
    \end{subfigure}
    \hspace{45pt}
    \begin{subfigure}[t]{0.41\linewidth}
        \centering
        \includegraphics[width=\linewidth]{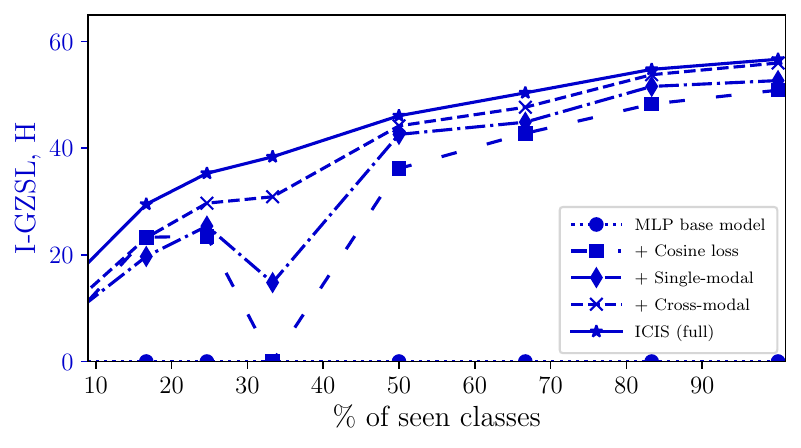}
        %\vspace{-2.0em}
        \caption{Impact on I-GZSL performance, harmonic mean H.}
        \label{fig:ablation_ours_numsamples_gzsl}
    \end{subfigure}
\caption{Data efficiency of architecture ablations of \ours\ on the CUB dataset. Over the various dataset sizes, each addition to the MLP base model up to the full \ours\ results in increasing performance for both the zero-shot and generalised zero-shot task. }
\vspace{-1.0em}
\label{fig:ablation_ours_numsamples_both}
\end{figure*}

\mypara{Data-scarcity and generalisation.}
Here, we further examine how the components of \ours\ combat overfitting and improve generalisation in this data-scarce setting. In \Cref{fig:ablation_ours_numsamples_both}, we show results for training our framework only with subsets of the descriptors of seen classes in the CUB dataset. We note that when training \ours\ on a lower number of descriptor-classifier pairs, we still evaluate the generalised zero-shot classification accuracy by injecting the predicted classifiers into the classification model with the full set of seen classes. 

We observe that when training \ours\ on dataset sizes varying from just 12 descriptor-classifier pairs to the full training set of 150 pairs, the performance of the classifiers predicted by \ours\ on both zero-shot classification and the generalised task increases consistently with each added component. In general,  \Cref{fig:ablation_ours_numsamples_both} confirms the trends of \Cref{table:ablation}, showing that each component improves the robustness to the number of training descriptor-classifier weights pairs in the extremely constrained \setting. Notably, when \ours\ has as little as 50\% of seen class descriptors, it already outperforms the MLP base model with full availability (\Cref{fig:ablation_ours_numsamples_both}). 

\mypara{Analysis of a failure case: \textit{Tree Sparrow}.}
In this section, we study some shortcomings of \ours\ on a concrete example considering the CUB dataset. In the generalised \setting,  on the unseen class \textit{Tree Sparrow} our model regresses classification weights which correctly classify only $5\%$ of the corresponding samples: this is the lowest accuracy our model achieves on an unseen class in this setting. We investigate this issue by counting which classes the \textit{Tree Sparrow} samples were misclassified with, ordering them by similarity with the target classes (as computed from their class descriptors). Since the majority of the predictions are located within the closest 10 classes (see supplementary), in \Cref{fig:failure} we show only these classes, ordered by similarity. 

As the figure shows, the injected classifiers from \ours\ result in classifying primarily the most similar classes to \textit{Tree Sparrow} - and mostly within the sparrow family. This suggests that the injected weights indeed classify based on properties close to those of \textit{Tree Sparrow}, showing that \ours\ captured its core visual properties. Among the top predicted classes, we can see that there are both seen and unseen ones, showing that our model is not overly biased toward seen classes. Indeed, misclassifications among seen classes only occurs on the two most similar classes to \textit{Tree Sparrow}, namely \textit{Chipping Sparrow} and \textit{House Sparrow}, that apart from extremely fine-grained details (\eg beak shape and size in \textit{House Sparrow}, see examples in the supplementary) are difficult to distinguish from the target class, in which cases the injected classifiers may fall short. Nevertheless, this analysis confirms that \ours\ has learned from the visual semantic knowledge encoded in existing classifier weights: this is reflected on the  injected weights for new, unseen, classes, being able to capture discriminative visual properties. 

\subsection{\setting\ from ImageNet}
\begin{figure}[t!]
    \centering
        \centering
        \includegraphics[width=\linewidth]{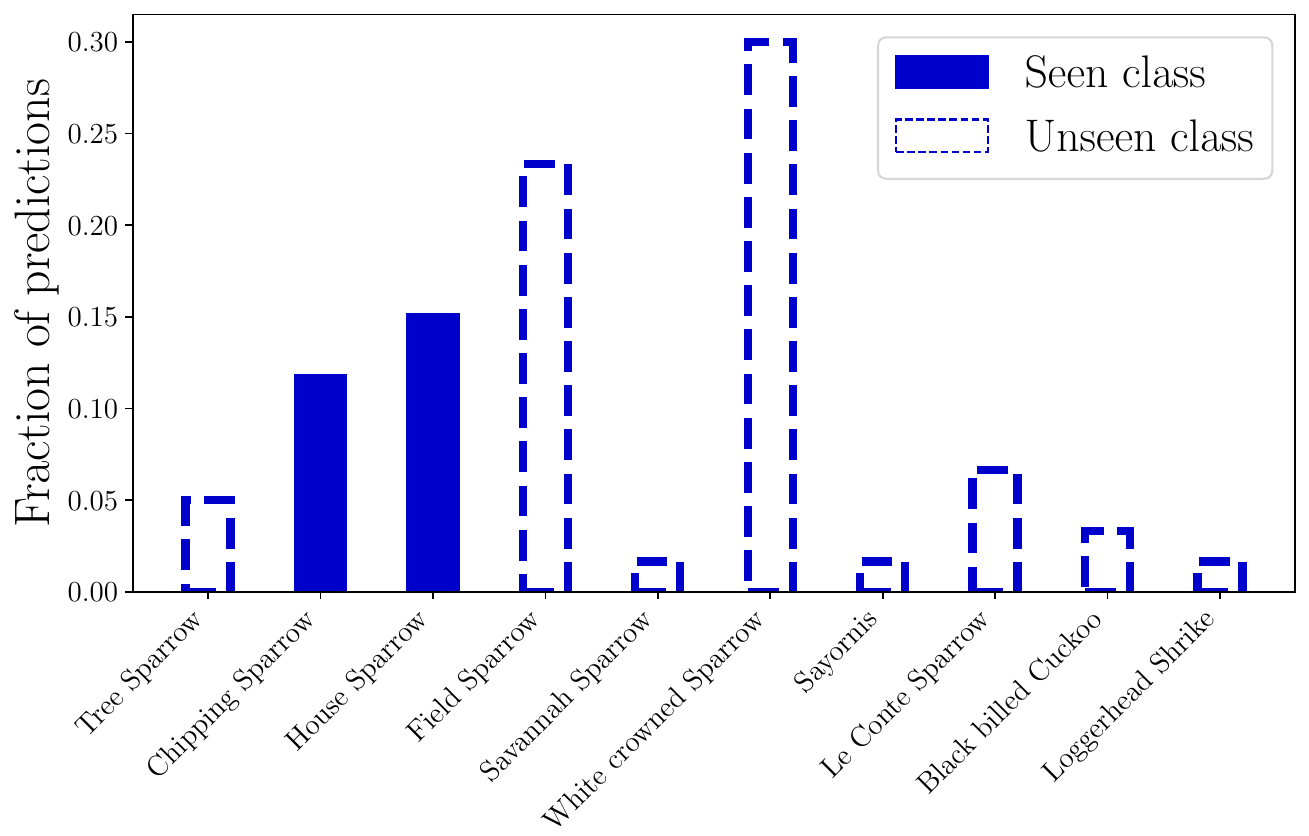}
        % \vspace{-2.0em}
        \caption{Predicted classes in the failure case \textit{Tree Sparrow} (CUB, I-GZSL), where the x-axis indicates predicted CUB classes ranked by attribute similarity to \textit{Tree Sparrow}, while the y-axis indicates the times an image of \textit{Tree Sparrow} is assigned to a class. The misclassifications are distributed mostly between classes of other unseen (dashed) Sparrow-classes, as well as the two seen classes (solid) that are closest to \textit{Tree Sparrow} in terms of attribute similarity.}
\label{fig:failure}
\end{figure}

In this section, we analyse how the ZS performance of \ours\ and the competitors change when i) \setting\ is performed from a generic pre-trained object classification model and ii) how the performance changes w.r.t. the available class descriptors. For the former, we consider a ResNet101~\cite{He2016DeepRL} pre-trained on ImageNet~\cite{deng2009imagenet}, since it is readily available online via libraries such as $\mathtt{torchvision}$ \cite{marcel2010torchvision}. For the embeddings, we consider Wikipedia word embeddings 
from \cite{yamada2020wikipedia2vec}, ConceptNet Numberbatch \cite{speer2017conceptnetemb} and CLIP text embeddings \cite{radford2021learning} that also encode visual knowledge. These embeddings have dimensionality 300, 300, and 512, respectively.

The results are shown in \Cref{table:zst_results}. \ours\ consistently achieves the top results across dataset and class descriptors, with the exception of Wiki2Vec and ConceptNet in CUB. These results, coupled with the ones in \Cref{table:results}, highlight the robustness of our method to the particular pre-trained classification model and the available embeddings, making \ours\ the most effective solution for \settingfull.  
 
From the table, there are also some general observations that can be drawn. Results are much lower than the I-ZSL ones in \Cref{table:results} for SUN and CUB in particular (\eg $-13.7\%$ and $-22.5\%$  for \ours\ with CLIP embeddings). Especially on CUB, the performances are very low for non-visually informed embeddings (\eg $6.0\%$ for wiki2vec, $10.2\%$ for ConceptNet). Since CUB is the dataset with the highest distribution shift w.r.t. ImageNet classes, these results highlight the challenges of \setting\ when seen and unseen classes largely differ. In such settings, non-trained approaches (\eg ConSE) can even outperform trained models (\eg \ours, wDAE) when class descriptors are less discriminative. On the other hand, on AWA2 the downstream performance of \ours\ when training it on a pre-trained ImageNet model can even outperform \ours\ on a model trained on the seen classes from AWA2 specifically, reaching zero-shot accuracies up to $86.1\%$. This is not surprising as the seen classes of AWA2 have a large overlap with the ImageNet classes, and that the number of descriptor-classifier pairs (and hence the training dataset for \ours) is 25 times larger (\ie from 40 to 1000). 
We note that the unseen target classes are disjoint from the ImageNet classes when using the split proposed in \cite{xianGoodBadUgly2}.
The results on AWA2 and SUN demonstrate that it is possible to approach \setting\ even when only a generic classifier is available, opening the possibility of constructing classification networks suited for tasks for which no visual dataset is available.

{\setlength{\tabcolsep}{2.5pt}
\renewcommand{\arraystretch}{1.1}
\begin{table}[t]
\centering
\resizebox{1.0\linewidth}{!}{
\begin{tabular}{l|ccc|ccc|ccc}
%\toprule
\multirow{2}{*}{\hspace{0.4cm}\textbf{Model}} &  \multicolumn{3}{c|}{\textbf{CUB}} & \multicolumn{3}{c|}{\textbf{AWA2}} &  \multicolumn{3}{c}{\textbf{SUN}} \\
& \bf WV & \bf CN & \bf CL & \bf WV & \bf CN & \bf CL & \bf WV & \bf CN & \bf CL  \\  \hline
ConSE~\cite{Norouzi2014ZeroShotLB} & $3.8$ & $\bf 11.6$ & $11.7$ & $50.3$ & $60.9$ & $72.9$ & $17.2$ & $25.4$ & $17.8$\\
COSTA~\cite{mensink2014costa}  & $3.9$ & $1.6$ & $14.1$ & $53.4$ & $51.8$ & $54.9$ & $12.5$ & $2.6$ & $18.8$\\
Sub.Reg.*~\cite{akyurek2021subspace}  & $3.7$ & $4.6$ & $4.5$ & $18.7$ & $60.0$ & $10.1$ & $3.7$ & $25.0$ & $1.6$\\
wDAE*~\cite{Gidaris2019GeneratingCW}  & $4.0$ & $6.0$ & $26.9$ & $64.4$ & $59.6$ & $84.5$ & $17.8$ & $24.6$ & $31.8$\\
WAvg*~\cite{xu2022vgse} & $5.5$ & $4.6$ & $12.2$ & $56.7$ & $58.8$ & $62.5$ & $11.9$ & $14.4$ & $6.5$\\
SMO*~\cite{xu2022vgse} & $\bf 7.3$ & $11.1$ & $12.7$ & $52.4$ & $64.5$ & $54.7$ & $16.4$ & $20.2$ & $13.1$\\
\ours\ (Ours) & $6.0$ & $10.2$ & $\bf 27.8$ & $\bf 65.2$ & $\bf 64.6$ & $\bf 86.1$ & $\bf 19.7$ & $\bf 25.9$ & $\bf 38.1$ \\
 %  \hline
\end{tabular}
}
\caption{Comparison between our \ours\ and existing methods in the image-free ZSL (I-ZSL) setting from ImageNet to the set of unseen classes of ZSL benchmarks. We also test various types of class descriptors, \ie Wiki2Vec~\cite{yamada2020wikipedia2vec} (WV), ConceptNet~\cite{speer2017conceptnetemb} (CN), and CLIP~\cite{radford2021learning} text embeddings (CL). We measure the results as unseen accuracy for the zero-shot task. Methods marked with *
are adapted to \setting.
}
\label{table:zst_results}

\end{table}
}

\section{Conclusion}
\label{sec:conclusion}
In this work, we presented the task of \settingfull\ (\setting) of pre-trained models which generalises zero-shot learning to a setting in which access to any per-sample information (such as images or image features) is not possible. We present a simple framework, \oursFull\ (\ours) that allows for adapting any given classifier by predicting new classifier weights from semantic per-class descriptors (\ie attributes, word embeddings). \ours\ improves generalisation by regularising the mapping from descriptors to weights with additional mappings within and across the corresponding spaces, injecting semantic priors in the model. Our framework is computationally efficient and works under minimal assumptions and supervision, \ie only classifier weights of the seen classes and their respective class descriptors. Our simple approach surpasses by a large margin all competitors, especially in the challenging generalised zero-shot classification task. \ours\ consistently outperforms competitors adapted from ZSL and FSL literature, demonstrating that the extremely constrained \settingfull\ for pre-trained models can be tackled effectively, opening avenues for future research in this topic. 
  
\newpage
\noindent\textbf{Acknowledgements.} This work was supported by DFG project number 276693517, BMBF FKZ: 01IS18039A, ERC (853489 - DEXIM), EXC number 2064/1 – project number 390727645, and by the MUR PNRR project FAIR - Future AI Research (PE00000013) funded by the NextGenerationEU. Ole Winther is supported by the Novo Nordisk Foundation (NNF20OC0062606) and the Pioneer Centre for AI, DNRF grant number P1. Anders Christensen thanks the ELLIS PhD program for support. 

{\small
\bibliographystyle{ieee_fullname}
\bibliography{references}

\begin{thebibliography}{10}\itemsep=-1pt

\bibitem{Akata2016LabelEmbeddingFI}
Zeynep Akata, Florent Perronnin, Za{\"i}d Harchaoui, and Cordelia Schmid.
\newblock Label-embedding for image classification.
\newblock {\em IEEE PAMI}, 2016.

\bibitem{akyurek2021subspace}
Afra~Feyza Aky{\"u}rek, Ekin Aky{\"u}rek, Derry Wijaya, and Jacob Andreas.
\newblock Subspace regularizers for few-shot class incremental learning.
\newblock In {\em ICLR}, 2021.

\bibitem{brown2020language}
Tom Brown, Benjamin Mann, Nick Ryder, Melanie Subbiah, Jared~D Kaplan, Prafulla
  Dhariwal, Arvind Neelakantan, Pranav Shyam, Girish Sastry, Amanda Askell,
  et~al.
\newblock Language models are few-shot learners.
\newblock In {\em NeurIPS}, 2020.

\bibitem{changpinyo2016synthesized}
Soravit Changpinyo, Wei-Lun Chao, Boqing Gong, and Fei Sha.
\newblock Synthesized classifiers for zero-shot learning.
\newblock In {\em CVPR}, 2016.

\bibitem{chao2016empirical}
Wei-Lun Chao, Soravit Changpinyo, Boqing Gong, and Fei Sha.
\newblock An empirical study and analysis of generalized zero-shot learning for
  object recognition in the wild.
\newblock In {\em ECCV}, 2016.

\bibitem{chen2022msdn}
Shiming Chen, Ziming Hong, Guo-Sen Xie, Wenhan Yang, Qinmu Peng, Kai Wang, Jian
  Zhao, and Xinge You.
\newblock Msdn: Mutually semantic distillation network for zero-shot learning.
\newblock In {\em CVPR}, 2022.

\bibitem{chen2021free}
Shiming Chen, Wenjie Wang, Beihao Xia, Qinmu Peng, Xinge You, Feng Zheng, and
  Ling Shao.
\newblock Free: Feature refinement for generalized zero-shot learning.
\newblock In {\em ICCV}, 2021.

\bibitem{cheraghian2021semanticfsl}
Ali Cheraghian, Shafin Rahman, Pengfei Fang, Soumava~Kumar Roy, Lars Petersson,
  and Mehrtash Harandi.
\newblock Semantic-aware knowledge distillation for few-shot class-incremental
  learning.
\newblock In {\em CVPR}, 2021.

\bibitem{chou2020adaptive}
Yu-Ying Chou, Hsuan-Tien Lin, and Tyng-Luh Liu.
\newblock Adaptive and generative zero-shot learning.
\newblock In {\em ICLR}, 2020.

\bibitem{de2021continual}
Matthias De~Lange, Rahaf Aljundi, Marc Masana, Sarah Parisot, Xu Jia,
  Ale{\v{s}} Leonardis, Gregory Slabaugh, and Tinne Tuytelaars.
\newblock A continual learning survey: Defying forgetting in classification
  tasks.
\newblock {\em IEEE PAMI}, 2021.

\bibitem{deng2009imagenet}
Jia Deng, Wei Dong, Richard Socher, Li-Jia Li, Kai Li, and Li Fei-Fei.
\newblock Imagenet: A large-scale hierarchical image database.
\newblock In {\em CVPR}, 2009.

\bibitem{dong2021few}
Songlin Dong, Xiaopeng Hong, Xiaoyu Tao, Xinyuan Chang, Xing Wei, and Yihong
  Gong.
\newblock Few-shot class-incremental learning via relation knowledge
  distillation.
\newblock In {\em AAAI}, 2021.

\bibitem{douillard2020podnet}
Arthur Douillard, Matthieu Cord, Charles Ollion, Thomas Robert, and Eduardo
  Valle.
\newblock Podnet: Pooled outputs distillation for small-tasks incremental
  learning.
\newblock In {\em ECCV}, 2020.

\bibitem{ferjad2022i2mvformer}
Muhammad Ferjad~Naeem, Muhammad Gul Zain~Ali Khan, Yongqin Xian, Muhammad
  Zeshan~Afzal, Didier Stricker, Luc Van~Gool, and Federico Tombari.
\newblock I2mvformer: Large language model generated multi-view document
  supervision for zero-shot image classification.
\newblock In {\em CVPR}, 2023.

\bibitem{french1999catastrophic}
Robert~M French.
\newblock Catastrophic forgetting in connectionist networks.
\newblock {\em Trends in cognitive sciences}, 3(4):128--135, 1999.

\bibitem{Gidaris2019GeneratingCW}
Spyros Gidaris and Nikos Komodakis.
\newblock Generating classification weights with gnn denoising autoencoders for
  few-shot learning.
\newblock {\em CVPR}, 2019.

\bibitem{HyperNetworksHa}
David Ha, Andrew~M. Dai, and Quoc~V. Le.
\newblock Hypernetworks.
\newblock In {\em 5th International Conference on Learning Representations,
  {ICLR} 2017, Toulon, France, April 24-26, 2017, Conference Track
  Proceedings}, 2017.

\bibitem{han2021contrastive}
Zongyan Han, Zhenyong Fu, Shuo Chen, and Jian Yang.
\newblock Contrastive embedding for generalized zero-shot learning.
\newblock In {\em CVPR}, 2021.

\bibitem{He2016DeepRL}
Kaiming He, X. Zhang, Shaoqing Ren, and Jian Sun.
\newblock Deep residual learning for image recognition.
\newblock {\em CVPR}, 2016.

\bibitem{hersche2022constrained}
Michael Hersche, Geethan Karunaratne, Giovanni Cherubini, Luca Benini, Abu
  Sebastian, and Abbas Rahimi.
\newblock Constrained few-shot class-incremental learning.
\newblock In {\em CVPR}, 2022.

\bibitem{kampffmeyer2019rethinking}
Michael Kampffmeyer, Yinbo Chen, Xiaodan Liang, Hao Wang, Yujia Zhang, and
  Eric~P Xing.
\newblock Rethinking knowledge graph propagation for zero-shot learning.
\newblock In {\em CVPR}, 2019.

\bibitem{kankuekul2012online}
Pichai Kankuekul, Aram Kawewong, Sirinart Tangruamsub, and Osamu Hasegawa.
\newblock Online incremental attribute-based zero-shot learning.
\newblock In {\em CVPR}, 2012.

\bibitem{keshari2020generalized}
Rohit Keshari, Richa Singh, and Mayank Vatsa.
\newblock Generalized zero-shot learning via over-complete distribution.
\newblock In {\em CVPR}, 2020.

\bibitem{kingma2014adam}
Diederik~P Kingma and Jimmy Ba.
\newblock Adam: A method for stochastic optimization.
\newblock In {\em ICLR}, 2015.

\bibitem{kipf2017semi}
Thomas~N Kipf and Max Welling.
\newblock Semi-supervised classification with graph convolutional networks.
\newblock In {\em ICLR}, 2017.

\bibitem{Kirkpatrick2017OvercomingCF}
James Kirkpatrick, Razvan Pascanu, Neil~C. Rabinowitz, Joel Veness, Guillaume
  Desjardins, Andrei~A. Rusu, Kieran Milan, John Quan, Tiago Ramalho, Agnieszka
  Grabska-Barwinska, Demis Hassabis, Claudia Clopath, Dharshan Kumaran, and
  Raia Hadsell.
\newblock Overcoming catastrophic forgetting in neural networks.
\newblock {\em Proceedings of the National Academy of Sciences}, 2017.

\bibitem{kong2022compactness}
Xia Kong, Zuodong Gao, Xiaofan Li, Ming Hong, Jun Liu, Chengjie Wang, Yuan Xie,
  and Yanyun Qu.
\newblock En-compactness: Self-distillation embedding \& contrastive generation
  for generalized zero-shot learning.
\newblock In {\em CVPR}, 2022.

\bibitem{LTDUO}
Christoph~H. Lampert, Hannes Nickisch, and Stefan Harmeling.
\newblock Learning to detect unseen object classes by between-class attribute
  transfer.
\newblock In {\em CVPR}, 2009.

\bibitem{lampert2013attribute}
Christoph~H Lampert, Hannes Nickisch, and Stefan Harmeling.
\newblock Attribute-based classification for zero-shot visual object
  categorization.
\newblock {\em IEEE PAMI}, 2013.

\bibitem{Li2019RethinkingZL}
K. Li, Martin~Renqiang Min, and Yun~Raymond Fu.
\newblock Rethinking zero-shot learning: A conditional visual classification
  perspective.
\newblock In {\em ICCV}, 2019.

\bibitem{li2017learning}
Zhizhong Li and Derek Hoiem.
\newblock Learning without forgetting.
\newblock {\em IEEE PAMI}, 2017.

\bibitem{GZSL-CalNet}
Shichen Liu, Mingsheng Long, Jianmin Wang, and Michael~I Jordan.
\newblock Generalized zero-shot learning with deep calibration network.
\newblock In S. Bengio, H. Wallach, H. Larochelle, K. Grauman, N. Cesa-Bianchi,
  and R. Garnett, editors, {\em Advances in Neural Information Processing
  Systems}, volume~31. Curran Associates, Inc., 2018.

\bibitem{lopez2017gradient}
David Lopez-Paz and Marc'Aurelio Ranzato.
\newblock Gradient episodic memory for continual learning.
\newblock {\em NeurIPS}, 2017.

\bibitem{Mahsereci2017EarlySW}
Maren Mahsereci, Lukas Balles, Christoph Lassner, and Philipp Hennig.
\newblock Early stopping without a validation set.
\newblock {\em ArXiv}, abs/1703.09580, 2017.

\bibitem{marcel2010torchvision}
S{\'e}bastien Marcel and Yann Rodriguez.
\newblock Torchvision the machine-vision package of torch.
\newblock In {\em ACM MM}, 2010.

\bibitem{mensink2014costa}
Thomas Mensink, Efstratios Gavves, and Cees~GM Snoek.
\newblock Costa: Co-occurrence statistics for zero-shot classification.
\newblock In {\em CVPR}, 2014.

\bibitem{min2020domain}
Shaobo Min, Hantao Yao, Hongtao Xie, Chaoqun Wang, Zheng-Jun Zha, and Yongdong
  Zhang.
\newblock Domain-aware visual bias eliminating for generalized zero-shot
  learning.
\newblock In {\em CVPR}, 2020.

\bibitem{naeem2022i2dformer}
Muhammad~Ferjad Naeem, Yongqin Xian, Luc Van~Gool, and Federico Tombari.
\newblock I2dformer: Learning image to document attention for zero-shot image
  classification.
\newblock In {\em NeurIPS}, 2022.

\bibitem{Norouzi2014ZeroShotLB}
Mohammad Norouzi, Tomas Mikolov, Samy Bengio, Yoram Singer, Jonathon Shlens,
  Andrea Frome, Gregory~S. Corrado, and Jeffrey Dean.
\newblock Zero-shot learning by convex combination of semantic embeddings.
\newblock In {\em ICLR}, 2014.

\bibitem{Parisot2023LearningTN}
Sarah Parisot, Yongxin Yang, and Steven~G. McDonagh.
\newblock Learning to name classes for vision and language models.
\newblock {\em ArXiv}, abs/2304.01830, 2023.

\bibitem{SUNdata}
Genevieve Patterson, Chen Xu, Hang Su, and James Hays.
\newblock The sun attribute database: Beyond categories for deeper scene
  understanding.
\newblock {\em IJCV}, 2014.

\bibitem{prabhu2020gdumb}
Ameya Prabhu, Philip~HS Torr, and Puneet~K Dokania.
\newblock Gdumb: A simple approach that questions our progress in continual
  learning.
\newblock In {\em ECCV}, 2020.

\bibitem{radford2021learning}
Alec Radford, Jong~Wook Kim, Chris Hallacy, Aditya Ramesh, Gabriel Goh,
  Sandhini Agarwal, Girish Sastry, Amanda Askell, Pamela Mishkin, Jack Clark,
  et~al.
\newblock Learning transferable visual models from natural language
  supervision.
\newblock In {\em ICML}, 2021.

\bibitem{rebuffi2017icarl}
Sylvestre-Alvise Rebuffi, Alexander Kolesnikov, Georg Sperl, and Christoph~H
  Lampert.
\newblock icarl: Incremental classifier and representation learning.
\newblock In {\em CVPR}, 2017.

\bibitem{romera2015embarrassingly}
Bernardino Romera-Paredes and Philip Torr.
\newblock An embarrassingly simple approach to zero-shot learning.
\newblock In {\em ICML}, 2015.

\bibitem{CADA-VAE}
Edgar Schönfeld, Sayna Ebrahimi, Samarth Sinha, Trevor Darrell, and Zeynep
  Akata.
\newblock Generalized zero- and few-shot learning via aligned variational
  autoencoders.
\newblock In {\em cvpr}, 2019.

\bibitem{speer2017conceptnetemb}
Robyn Speer and Joanna Lowry-Duda.
\newblock Conceptnet at semeval-2017 task 2: Extending word embeddings with
  multilingual relational knowledge.
\newblock In {\em International Workshop on Semantic Evaluation
  (SemEval-2017)}, 2017.

\bibitem{tao2020few}
Xiaoyu Tao, Xiaopeng Hong, Xinyuan Chang, Songlin Dong, Xing Wei, and Yihong
  Gong.
\newblock Few-shot class-incremental learning.
\newblock In {\em CVPR}, 2020.

\bibitem{CUBdata}
Catherine Wah, Steve Branson, Peter Welinder, Pietro Perona, and Serge~J.
  Belongie.
\newblock The caltech-ucsd birds-200-2011 dataset.
\newblock 2011.

\bibitem{wang2018zero}
Xiaolong Wang, Yufei Ye, and Abhinav Gupta.
\newblock Zero-shot recognition via semantic embeddings and knowledge graphs.
\newblock {\em CVPR}, 2018.

\bibitem{wei2021incremental2}
Kun Wei, Cheng Deng, Xu Yang, and Maosen Li.
\newblock Incremental embedding learning via zero-shot translation.
\newblock In {\em AAAI}, 2021.

\bibitem{wei2021incremental1}
Kun Wei, Cheng Deng, Xu Yang, and Dacheng Tao.
\newblock Incremental zero-shot learning.
\newblock {\em IEEE Transactions on Cybernetics}, 2021.

\bibitem{xian2016latent}
Yongqin Xian, Zeynep Akata, Gaurav Sharma, Quynh Nguyen, Matthias Hein, and
  Bernt Schiele.
\newblock Latent embeddings for zero-shot classification.
\newblock In {\em CVPR}, 2016.

\bibitem{xian2019semantic}
Yongqin Xian, Subhabrata Choudhury, Yang He, Bernt Schiele, and Zeynep Akata.
\newblock Semantic projection network for zero-and few-label semantic
  segmentation.
\newblock In {\em CVPR}, 2019.

\bibitem{xianGoodBadUgly2}
Yongqin Xian, H.~Christoph Lampert, Bernt Schiele, and Zeynep Akata.
\newblock Zero-shot learning - a comprehensive evaluation of the good, the bad
  and the ugly.
\newblock {\em IEEE PAMI}, 2018.

\bibitem{Xian2018FeatureGN}
Yongqin Xian, Tobias Lorenz, Bernt Schiele, and Zeynep Akata.
\newblock Feature generating networks for zero-shot learning.
\newblock {\em CVPR}, 2018.

\bibitem{xianGoodBadUgly1}
Yongqin Xian, Bernt Schiele, and Zeynep Akata.
\newblock Zero-shot learning - the good, the bad and the ugly.
\newblock In {\em CVPR}, 2017.

\bibitem{XianCVPR2019a}
Yongqin Xian, Saurabh Sharma, Bernt Schiele, and Zeynep Akata.
\newblock f-vaegan-d2: A feature generating framework for any-shot learning.
\newblock In {\em CVPR}, 2019.

\bibitem{xie2020region}
Guo-Sen Xie, Li Liu, Fan Zhu, Fang Zhao, Zheng Zhang, Yazhou Yao, Jie Qin, and
  Ling Shao.
\newblock Region graph embedding network for zero-shot learning.
\newblock In {\em ECCV}, 2020.

\bibitem{xu2020attribute}
Wenjia Xu, Yongqin Xian, Jiuniu Wang, Bernt Schiele, and Zeynep Akata.
\newblock Attribute prototype network for zero-shot learning.
\newblock In {\em NeurIPS}, 2020.

\bibitem{xu2022vgse}
Wenjia Xu, Yongqin Xian, Jiuniu Wang, Bernt Schiele, and Zeynep Akata.
\newblock Vgse: Visually-grounded semantic embeddings for zero-shot learning.
\newblock In {\em CVPR}, 2022.

\bibitem{xue2017incremental}
Nan Xue, Yi Wang, Xin Fan, and Maomao Min.
\newblock Incremental zero-shot learning based on attributes for image
  classification.
\newblock In {\em ICIP}, 2017.

\bibitem{yamada2020wikipedia2vec}
Ikuya Yamada, Akari Asai, Jin Sakuma, Hiroyuki Shindo, Hideaki Takeda,
  Yoshiyasu Takefuji, and Yuji Matsumoto.
\newblock Wikipedia2vec: An efficient toolkit for learning and visualizing the
  embeddings of words and entities from wikipedia.
\newblock In {\em EMNLP}, 2020.

\bibitem{zenke2017continual}
Friedemann Zenke, Ben Poole, and Surya Ganguli.
\newblock Continual learning through synaptic intelligence.
\newblock In {\em ICML}, 2017.

\bibitem{Zhang2021FewShotIL}
Chi Zhang, Nan Song, Guosheng Lin, Yun Zheng, Pan Pan, and Yinghui Xu.
\newblock Few-shot incremental learning with continually evolved classifiers.
\newblock {\em CVPR}, 2021.

\bibitem{Zhang2017LearningAD}
Li Zhang, Tao Xiang, and Shaogang Gong.
\newblock Learning a deep embedding model for zero-shot learning.
\newblock {\em CVPR}, 2017.

\bibitem{Zhu2018AGA}
Yizhe Zhu, Mohamed Elhoseiny, Bingchen Liu, Xi Peng, and A. Elgammal.
\newblock A generative adversarial approach for zero-shot learning from noisy
  texts.
\newblock {\em CVPR}, 2018.

\bibitem{Zhu2019LearningFT}
Yizhe Zhu, Jianwen Xie, Bingchen Liu, and A. Elgammal.
\newblock Learning feature-to-feature translator by alternating
  back-propagation for generative zero-shot learning.
\newblock {\em ICCV}, 2019.

\end{thebibliography}
}

\clearpage 
\appendix
\begin{center}
\large{\textbf{Supplementary Material:\\ Image-free Classifier Injection for Zero-Shot Classification}}
\end{center}
In this supplementary material, we firstly in Sec.~\ref{sec:implementation-details} provide additional implementation details regarding the baselines and our stopping criterion. In Sec.~\ref{sec:improving_baselines}, we combine applicable baselines with \ours\ and report the combined performance. We report additional results for the I-GZSL task for an ImageNet pretrained classification model in Sec.~\ref{section:additional_results_zst}.
In Sec.~\ref{section:seen_cls_bias}, we show that the architectural additions of \ours\ deal with bias towards seen classes, a common issue of GZSL, despite having no access to images. 
We finally provide an extended analysis of a failure case of our model in the generalized zero-shot task on CUB in Sec.~\ref{sec:failures}.

\section{Additional implementation details}
\label{sec:implementation-details}

\mypara{Implementation and adaptation of baselines.}
In this section, we detail how the baselines used for comparison with \ours\ have been implemented (and potentially adapted) for the \settingfull\ task.

\textit{ConSE}: We follow the original implementation of \cite{Norouzi2014ZeroShotLB}, using the classifier scores to perform a convex combination of the class label embeddings. 

\textit{COSTA}: In \cite{mensink2014costa},  different co-occurrence similarities are proposed. In our experiments, normalised co-occurrence similarity led to the best results.  

\textit{Sub. Reg.*}: We adapt the subspace regularisers proposed in \cite{akyurek2021subspace} originally acting on model parameters, to instead serve as an additional loss on the predicted classifiers during training. Concretely, we apply a projection matrix $P_S$ on the predicted classifiers, and apply a loss based on the distance $d$ between the predicted and projected weights:
\begin{equation}
    \label{eq:subspace_reg}
     \mathcal{L}_{Sub. Reg.} = \sum_{u\in U} d(\mathbf{\cls}_u, P^T_S\mathbf{\cls}_u).
\end{equation}
The projection matrix $P_S$ is calculated from the existing classifiers of the seen classes. It projects a predicted weight $\mathbf{\cls}_u$ onto the subspace spanned by the existing seen class classifiers. We found that using the squared error, as originally done in \cite{akyurek2021subspace}, worked the best.

\textit{wDAE*}: In order to adapt the setup from \cite{Gidaris2019GeneratingCW} to the \setting\ setting, two primary changes were made. Firstly, since image features are not available, classifier estimates originally created by feature averaging are not accessible. Therefore, we train a weight predictor to provide the initial estimates. In the main paper, we use the MLP base model as the initial predictor, but here in the supplementary material we also report results when used in conjunction with \ours\ (\Cref{table:extended_results}).
Secondly, for the proposed denoising autoencoder (DAE) to be applicable in our setting, we adapt the loss function for the model. Since we do not have access to images to test downstram classification accuracy, we remove the image-related term of Eq. (6). in \cite{Gidaris2019GeneratingCW}, resulting in the loss function
\begin{equation}
    \label{eq:wDAE}
     \mathcal{L}_{wDAE} = \sum_{u\in U} d(\mathbf{\cls}_u, \mathbf{\cls}_u^*).
\end{equation}
Here, $\mathbf{\cls}_u$ are the initially estimated weights, and $\mathbf{\cls}_u^*$ are the reconstructed weights by the DAE, with $d$ as mean squared distance leading to the best downstream results.

\textit{VGSE (WAvg* and SMO*}): We adapt the weighting schemes of \cite{xu2022vgse} to directly produce new classifiers. For WAvg, we use the same hyperparameters as presented in the paper. For SMO, we experimented with both $\alpha=0$ and $\alpha=-1$. However, $\alpha=0$ consistenly led to better downstream results, and all reported results are computed with this value.

\mypara{Stopping criterion.} We take inspiration from \cite{Mahsereci2017EarlySW}, where an early stopping criterion based on statistical properties of the gradients was proposed. Here, we base our simple stopping criterion on the slope of the training loss. Concretely, we compute a running mean of the training loss over the latest 10 epochs, as well as the 10 previous epochs. We then compute the (negative) slope between these averages, and compare it to a threshold value which is set to $2\cdot 10^{-4}$. This value was determined from the simple empirical observations that higher values were too high (\ie training was terminated almost immediately across different architectures), while lower values were too low (\ie the slope rarely went below the threshold, such that the training loop continued indefinitely, despite the network already having converged). When applying the mean squared error as the loss function (instead of the cosine loss), the losses are smaller by a factor of approximately $10^{-3}$ and the threshold value is therefore then scaled by this factor.

\section{Improving baselines with \ours}
\label{sec:improving_baselines}
As the model components that ultimately make up \ours\ are orthogonal to Sub. Reg.* and wDAE*, we include results using these baselines in combination with our \ours\ framework in \Cref{table:extended_results}. For the results reported under Sub.~Reg.* and wDAE*, the initial weight estimator network is an MLP network with a similar structure as the descriptor-to-weights mapping of \ours\ (\ie the MLP base model performance from our ablation study). For this initial weight predictor, Sub.~Reg.* and wDAE* do not improve results, compared to the results of the MLP base model by itself. Indeed, the zero-shot accuracies generally drop (\eg $41.3\%$ to $37.6\%$ on CUB for Sub. Reg.*, as can be seen in Table 1 and Table 2 of the paper), and the harmonic mean does not increase.

However, if we apply \ours\ to wDAE* and Sub. Reg* together, we see consistent performance gains across all datasets and setting. For instance, the zero-shot accuracy increases of more than 20\% on both AWA2 and CUB for both methods, while on SUN the performance gain is at least of 2\%. The gap is even clearer in the generalised setting, where Sub. Reg.* and wDAE* are able to correctly recognise unseen classes only when coupled with \ours, going from 0 harmonic mean to over 50 for CUB and AWA2, while of more than 32 for SUN. These results provide further evidence of how \ours\ reduces the bias on seen classes in image-free generalised zero-shot learning. Furthermore, they demonstrate the flexibility of \ours, being a simple approach that can benefit the performance of other methods based on classifier weights prediction. 

{\setlength{\tabcolsep}{4pt}
\renewcommand{\arraystretch}{1.1}
\begin{table*}[t]
\centering
\resizebox{0.9\linewidth}{!}{
\begin{tabular}{l | c |ccc|ccccccccc}
&                                                   & \multicolumn{3}{c|}{\textbf{Zero-Shot Accuracy}}                                                                                            & \multicolumn{9}{c}{\textbf{Generalised Zero-Shot Accuracy}}                                                                                                                                                                                                                                                                                                                                                                            \\
\multirow{2}{*}{\textbf{Method}} & \multirow{2}{*}{\textbf{+ ICIS}}                                                  & \textbf{CUB}                                 & \textbf{AWA2}                                & \textbf{SUN}                                  & \multicolumn{3}{c}{\textbf{CUB}}                                                                                                            & \multicolumn{3}{c}{\textbf{AWA2}}                                                                                                           & \multicolumn{3}{c}{\textbf{SUN}}                                                                                                           \\               
&   & \textbf{Acc}                                  & \textbf{Acc}                                  & \textbf{Acc}                                   & \textbf{u}                                   & \textbf{s}                                   & \multicolumn{1}{c|}{\textbf{H}}               & \textbf{u}                                   & \textbf{s}                                   & \multicolumn{1}{c|}{\textbf{H}}               & \textbf{u}                                   & \textbf{s}                                   & \textbf{H}                                   \\ \hline
\multirow{2}{*}{Sub.~Reg.* \cite{akyurek2021subspace}} &             \ding{55}                                                     &        $37.6$                                    &     $37.5$                                         &         $48.3$                                      &       $0.0$                                       &       $87.6$                                       & \multicolumn{1}{c|}{$0.0$}                         &                                    $0.0$          &                                 $96.1$             & \multicolumn{1}{c|}{$0.0$}                         &          $0.0$                                    &          $50.1$                                    &     $0.0$                                         \\
 & \ding{51}                                                                   & $\mathbf{60.5}$                                 &                $\mathbf{65.8}$                              & $\mathbf{51.8}$                                        &              $45.8$                                &   $73.6$                                           & \multicolumn{1}{c|}{$\mathbf{56.5}$}                         &                              $35.7$                &         $93.3$                                     & \multicolumn{1}{c|}{$\mathbf{51.6}$}                         &  $45.6$                                            &                $25.1$                              &   $\mathbf{32.3}$                                           \\
\hline 
\multirow{2}{*}{wDAE*  \cite{Gidaris2019GeneratingCW}}  &       \ding{55}                                              & $38.2$                                 &          $37.0$                                    &                                    $49.9$           & $0.0$                                  & $87.3$                                 & \multicolumn{1}{c|}{$0.0$}              &     $0.1$                                         &           $96.0$                                   &  \multicolumn{1}{c|}{$0.3$}                         &             $0.0$                                 &                   $49.3$                           &       $0.0$                                       \\  
 & \ding{51}                                                                   &$\mathbf{60.4}$                                &               $\mathbf{65.1}$                           & $\mathbf{51.9} $                                        &              $46.1$                                &   $73.1$                                           & \multicolumn{1}{c|}{$\mathbf{56.6}$}                         &                              $36.0 $                &         $93.3$                                     & \multicolumn{1}{c|}{$\mathbf{52.0}$}                         &  $45.1$                                            &                $25.5$                              &   $\mathbf{32.6}$                                           \\                     
\end{tabular}
}
\caption{Improving downstream classification performance on ZSL benchmarks in the \setting\ setting when combining existing methods in the literature with our proposed \ours\ framework. We measure the results as unseen accuracy (Acc) for the zero-shot task, unseen (u) and seen (s) accuracy and their harmonic mean (H) for the generalised zero-shot setting. Methods marked with * are adapted to the image-free setting. \vspace{5pt}
}
\label{table:extended_results}

\end{table*}
}

\section{Additional results}
\label{section:additional_results_zst}
Here, we show results demonstrating the performance of \ours\ for different pre-training setups of the backbone, and for the I-GZSL task on CUB for an ImageNet pretrained model with injected CUB classifiers. 

Firstly, we analyse the impact of varying the ResNet-101 backbone. In \Cref{table:results}, the backbone is fine-tuned on the seen classes of the ZSL datasets, as commonly done in the literature \cite{XianCVPR2019a}. Alternatively, the classification head can be trained while keeping the feature extractor frozen. We report results for this option in the left column of \Cref{table:additional_results} on CUB. We observe that also when using an ImageNet-pretrained feature extractor, \ours\ still perform significantly better than competing methods in terms of both zero-shot accuracy (Acc) and on the generalized I-ZSL task in terms of resulting harmonic mean (H). Overall, the trend of the results do not change between using a pre-trained and fine-tuned feature extractor. Curiously, the performance of some methods increase slightly when using the ImageNet-pretrained classifier (e.g. wDAE*~\cite{Gidaris2019GeneratingCW}  with Acc and H increasing from $38.2$\% and $0.0$ to $40.0\%$ and $4.1$, respectively.

Secondly, we report I-GZSL results corresponding to \Cref{table:zst_results}, i.e.\ the generalized zero-shot classification task for an ImageNet pretrained model for which we inject classifiers for classification on CUB in \textit{addition} to classifying ImageNet. The results are reported in the right column of \Cref{table:additional_results}. We can see that the results from learning to inject classifiers from a classifier trained on a semantically unrelated dataset (in this case ImageNet) leads to lower results than when learning from a semantically related one (e.g. CUB seen classes). However, \ours\ is still able to infer and inject classifiers for the generalized task that lead to non-trivial performance on this difficult task (unseen accuracy $11.6\%$) while maintaining an accuracy of $77.3\%$ on the seen classes (with the pre-trained ResNet model achieving a $77.4\%$ seen class accuracy before injecting additional classifiers).

{\setlength{\tabcolsep}{4pt} %2.5
\renewcommand{\arraystretch}{1.1}
\begin{table}[t]
\centering
\resizebox{0.9\linewidth}{!}{
\begin{tabular}{l|cc|ccc}
\multirow{2}{*}{\hspace{0.4cm}\textbf{Model}} &  \multicolumn{2}{c|}{\textbf{CUB to CUB}} & \multicolumn{3}{c}{\textbf{ImageNet to CUB}} \\
& $\textrm{Acc}$ & $\textrm{H}$ & s & u &  $\textrm{H}$ \\ 
\hline
ConSE~\cite{Norouzi2014ZeroShotLB} & $35.9$ & $0.9$ & $77.0$ & $0.0$ & $0.0$   \\
COSTA~\cite{mensink2014costa} & $21.8$ & $0.0$ & $75.2$ & $6.7$ & $12.3$   \\
Sub.Reg.*~\cite{akyurek2021subspace}  & $40.3$ & $4.7$ & $77.4$ & $0.0$ & $0.0$   \\
wDAE*~\cite{Gidaris2019GeneratingCW} & $40.0$ & $4.1$ & $77.3$ & $7.4$ & $13.5$  \\
WAvg*~\cite{xu2022vgse} & $2.0$ & $1.8$ & $77.4$ & $0.0$ & $0.0$   \\
SMO*~\cite{xu2022vgse} & $39.3$ & $32.2$ & $70.0$ & $2.5$ & $4.9$ \\ 
\ours\ (Ours) & $\bf 50.8$ & $\bf 41.2$ & $77.3$ & $11.6$ & $\bf 20.2$  \\
\end{tabular}
}
\caption{
Left: CUB I-ZSL (Acc) and I-GZSL (H) results using a frozen ResNet feature extractor pre-trained on ImageNet. Right: I-GZSL results from ImageNet to CUB using CLIP class label text embeddings. Methods marked with *
are adapted to \setting.}
\label{table:additional_results}
\vspace{-1.4em}
\end{table}
}

\section{Dealing with bias towards seen classes}
\label{section:seen_cls_bias}
For the generalised zero-shot classification problem, bias towards seen classes is one of the primary challenges to overcome. In the GZSL literature, a common techniques to address this obstacle is to use seen and unseen class experts \cite{chou2020adaptive}, or to introduce hyperparameters that reduce the confidence for seen classes \cite{chao2016empirical}. 

\begin{figure}[t]
    \centering
    \includegraphics[width=\linewidth]{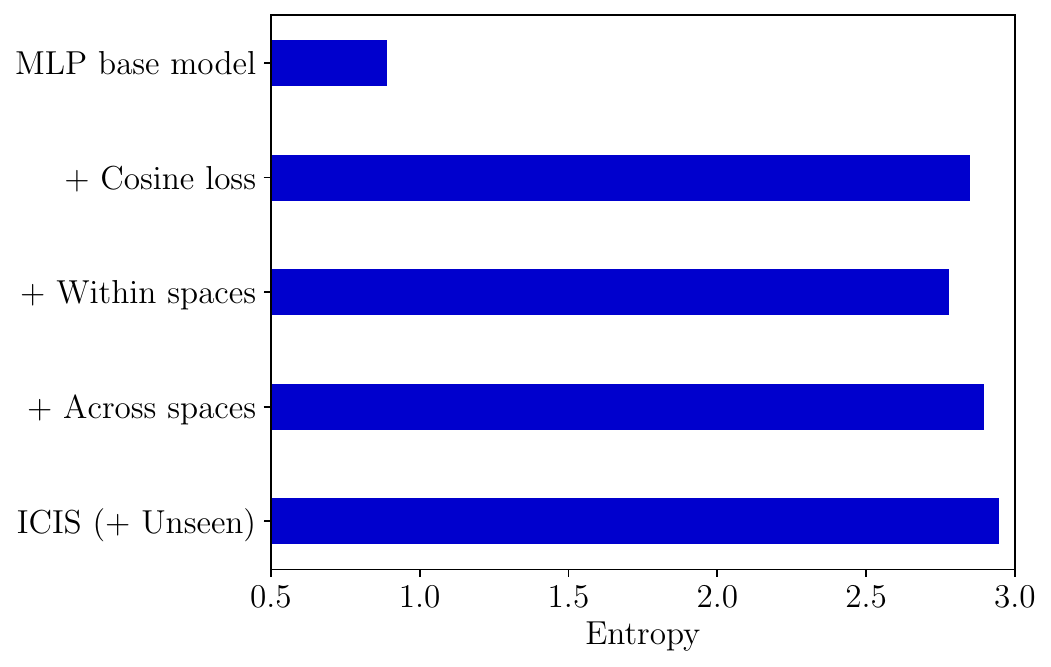}
    \vspace{-2.0em}
    \caption{Inference analysis of our model ablations on CUB. We plot the mean entropy (calculated using $\text{log}_e$) of the output distributions over the test samples from both seen and unseen classes. The cosine loss in particular has a large impact on the injected weights dealing with bias towards seen classes.
    }\vspace{-1.0em}
    \label{fig:entropy}
\end{figure}

In the image-free \setting\ setting, these options are not applicable, since there is no access to samples on which those options can be tuned. Instead, we show in the analysis in \Cref{fig:entropy} that each element of \ours\ increases the average entropy of the output distribution over the seen and unseen class test samples in CUB, \ie it reduces overconfidence towards seen classes. 

This analysis confirms the findings of the previous one in \Cref{table:extended_results}, as the baselines (without \ours) almost exclusively predict seen classes in the I-GZSL setting. 
However, the introduction of our framework's components increase the accuracy for unseen classes for the baselines notably, while having only a small impact on the seen class accuracy. 

\section{Study of failure case}
\label{sec:failures}
In this section, we study some shortcomings of \ours\ on a concrete example for the CUB dataset. Concretely, we consider the behaviour of the injected weights on the class for which they achieve the worst classification accuracy in the downstream I-GZSL task.  

\begin{figure*}[t]
    \centering
    \includegraphics[width=0.6\linewidth]{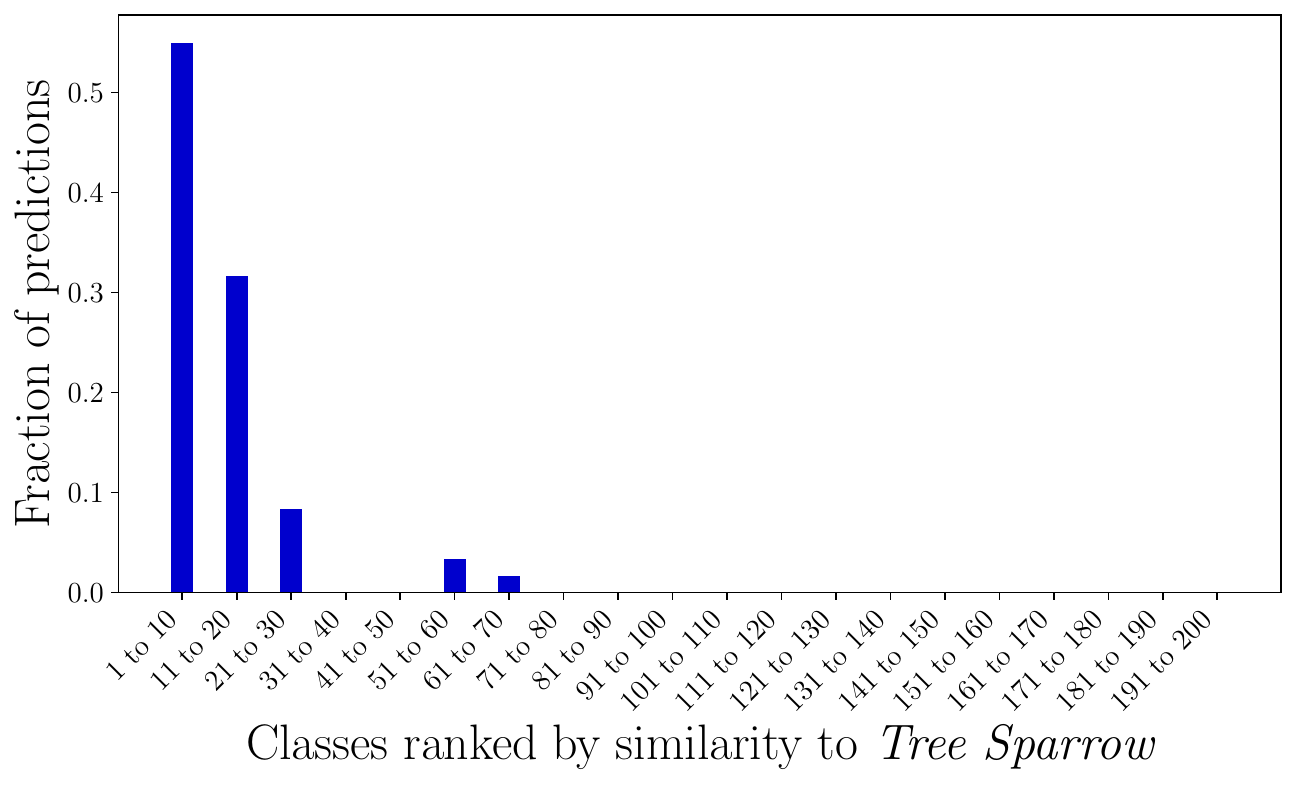}
    %\vspace{-2.0em}
    \caption{\textbf{Overview of the failure case \textit{Tree Sparrow}}, where the injected weights for the generalised zero-shot task on CUB only correctly classifies 5\% of the samples. The x-axis indicates CUB classes ranked by the attribute similarity to \textit{Tree Sparrow}, while the y-axis shows the number of times an image of the class \textit{Tree Sparrow} is assigned to a group of classes. Most misclassifications are towards classes similar to the ground-truth one. }
    %\vspace{-1.0em}
    \label{fig:tree_sparrow_bins}
\end{figure*}

\mypara{Failures of injected weights in the generalised task.}
For the unseen class \textit{Tree Sparrow}, our \ours\ injects classification weights which correctly classify only $5\%$ of the corresponding samples for the generalised zero-shot classification task. In Figure \ref{fig:tree_sparrow_bins}, we plot the predictions of our model for images of this class. 

For plotting the predictions, we count which classes were predicted for the \textit{Tree Sparrow} samples, and then divide these counts by the total number of \textit{Tree Sparrow} samples to obtain class-wise empirical (mis)classification probabilities. 
To get an understanding of the properties of the classes in the dataset that the \textit{Tree Sparrow} samples were misclassified as, we rank all 200 classes in the CUB dataset based on the cosine similarity between the attributes of \textit{Tree Sparrow}, and the attributes of each class in the dataset. For visualisation, we bin the classes in groups of 10, starting from the most similar (\ie \textit{Tree Sparrow} itself) to the least similar class in the dataset, summing up their empirical prediction probabilities. 

Figure \ref{fig:tree_sparrow_bins} shows how the predicted classification weights result in classifying primarily the most similar classes to \textit{Tree Sparrow}. This suggests that the predicted weights indeed classify based on properties close to those of \textit{Tree Sparrow}.

\begin{figure*}[t]
    \centering
    \includegraphics[width=0.6\linewidth]{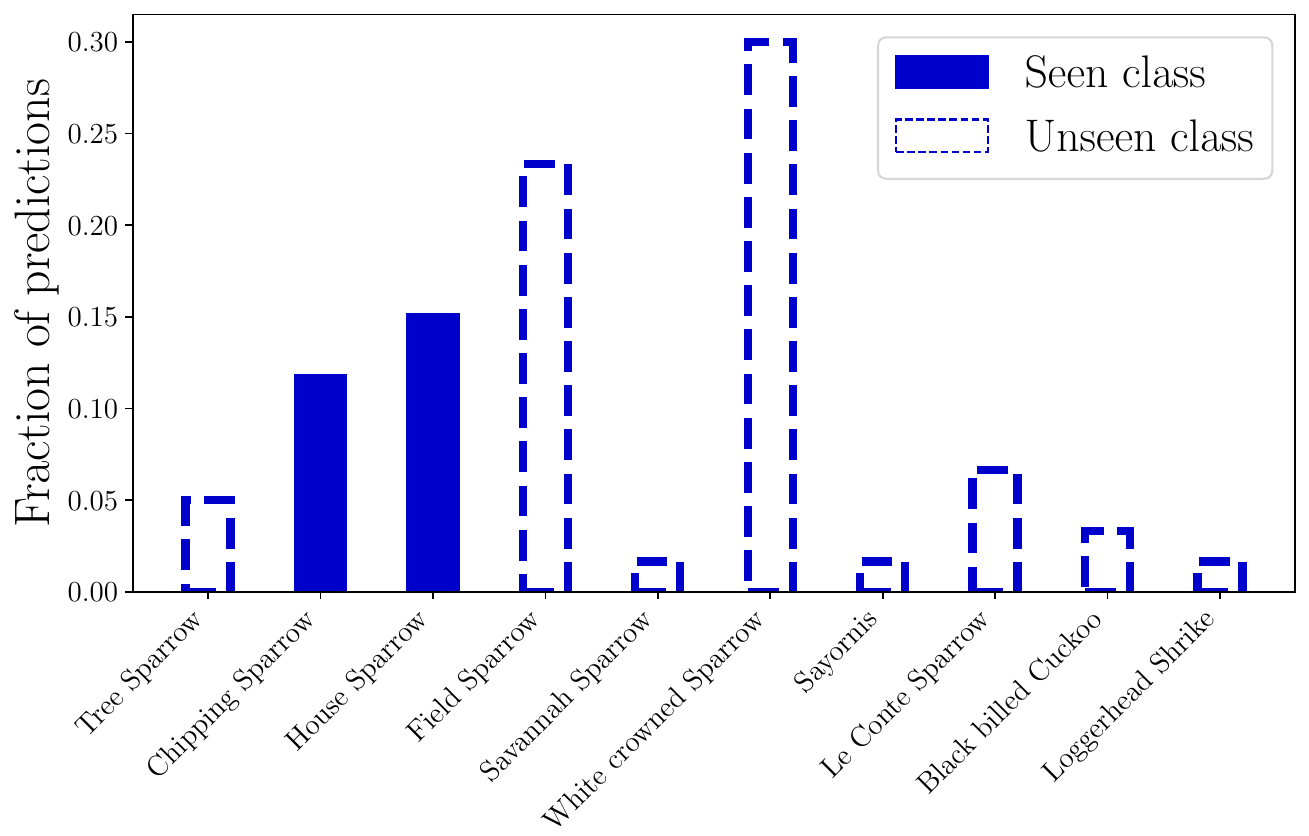}
    \caption{\textbf{Predicted classes in the failure case \textit{Tree Sparrow}}, where the x-axis indicates predicted CUB classes ranked by the attribute similarity to \textit{Tree Sparrow}, while the y-axis shows the frequency of an image of \textit{Tree Sparrow} being assigned to a group of classes. The misclassifications are distributed mostly between classes of other unseen (dashed) Sparrow-classes, as well as the two seen classes (solid) that are the closest to \textit{Tree Sparrow} in terms of attributes.}
    %\vspace{-1.0em}
    \label{fig:tree_sparrow_zoom}
\end{figure*}

Figure \ref{fig:tree_sparrow_zoom} zooms in on just the classes that the \textit{Tree Sparrow} samples were misclassified as, revealing that the predicted weights have a bias towards other unseen classes. Most misclassifications are due to predicting other unseen and similar classes, such as \textit{White-crowned Sparrow} and \textit{Field Sparrow}. Similarly, misclassifications among seen classes only occur on the two most similar classes to \textit{Tree Sparrow}, namely \textit{Chipping Sparrow} and \textit{House Sparrow}. 

\begin{figure*}[t]
    \centering
    \includegraphics[width=0.53\linewidth]{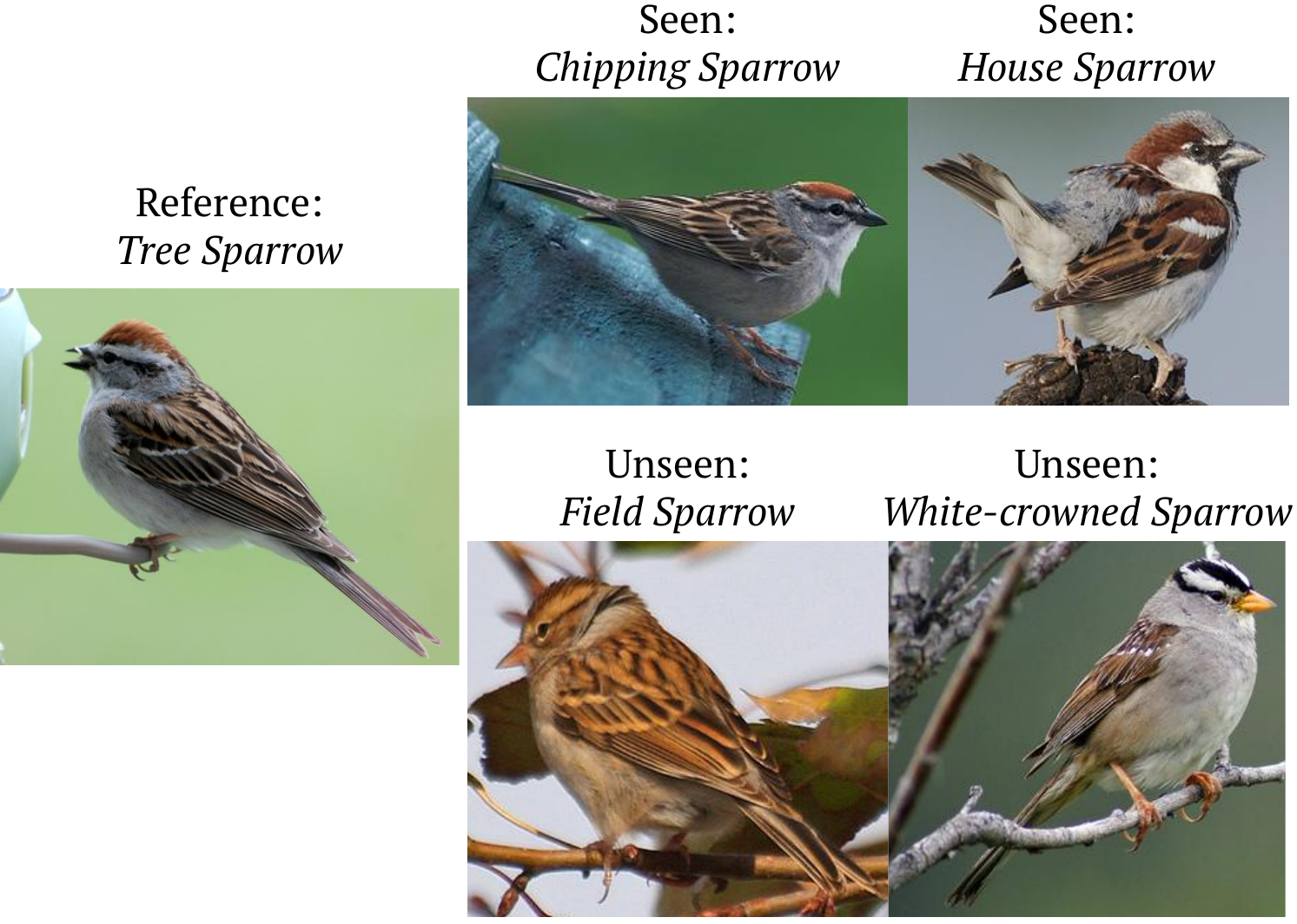}
    %\vspace{-2.0em}
    \caption{\textbf{Images from the seen and unseen classes most consistently confused with \textit{Tree Sparrow}} by the classification weights predicted by our \ours\ model.}
    %\vspace{-1.0em}
    \label{fig:sparrow_photos}
\end{figure*}

For reference, we show images of birds from the listed classes in Figure \ref{fig:sparrow_photos}. We include images from the two seen classes confused with Tree Sparrow (i.e. \textit{Chipping Sparrow} and \textit{House Sparrow}) as well as images from the two unseen classes with the largest fraction of predictions (i.e. \textit{White-crowned Sparrow} and \textit{Field Sparrow}). Although there are distinctive differences between classes (e.g. the black and white head of the \textit{White-crowned Sparrow}, or the beak shape and size of the \textit{House Sparrow}), it is clear that the predicted weights are capturing fundamental visual properties, but may not be able to capture extremely fine-grained differences among classes. Nevertheless, this indicates that \ours\ has indeed learned from the visual-semantic knowledge encoded in the given classifier weights for seen classes.

\end{document}